\documentclass{article}
\newif\ifarxivversion
\usepackage{iclr2027_conference,times}
\usepackage{microtype,graphicx,booktabs,longtable,array,amsmath,amssymb,amsthm,mathtools,xcolor,bm,enumitem,tikz}
\usepackage{hyperref,url}
\makeatletter
\g@addto@macro\UrlBreaks{\do\a\do\b\do\c\do\d\do\e\do\f\do\0\do\1\do\2\do\3\do\4\do\5\do\6\do\7\do\8\do\9}
\makeatother
\newcommand{\ebar}{\bar{e}}
\newcommand{\etil}{\tilde{e}}
\newcommand{\hb}{\bar{h}}
\newcommand{\kap}{\kappa}
\newcommand{\Hl}[1]{\|\bar h_{#1}\|^2}

\newcommand{\sech}{\operatorname{sech}}
\newcommand{\Cov}{\operatorname{Cov}}
\newcommand{\E}{\mathbb{E}}
\newcommand{\R}{\mathbb{R}}

\newtheorem{proposition}{Proposition}

\newcommand{\numPriorLoss}{3.2491}

\newcommand{\numHeadlineTraj}{15}

\newcommand{\numEbarInit}{1.278}

\newcommand{\numKappaThree}{4.6}

\newcommand{\numCenterEHL}{6.0}

\newcommand{\numBaseHLThree}{182}

\newcommand{\numCenterEEbarFifteen}{0.61}

\newcommand{\numBaseEbarFifteen}{0.05}

\newcommand{\numCenterELossFifteen}{3.35}

\newcommand{\numBaseLossFifteen}{1.31}

\newcommand{\numModelConds}{48}

\newcommand{\numCrossSeedRuns}{144}

\newcommand{\numGridR}{0.96}

\newcommand{\numGridMAE}{0.07}

\newcommand{\numGridExitPairs}{39}

\newcommand{\numGridExitError}{0.24}

\newcommand{\numGridMissedPlateaus}{4}

\newcommand{\numGridExtraPlateaus}{3}

\newcommand{\numProductLawSlope}{-1.75}

\newcommand{\numProductLawPrefactor}{0.0021}

\newcommand{\numProductLawR}{-0.989}

\newcommand{\numMasterN}{33}

\newcommand{\numMasterRange}{361-fold}

\newcommand{\numKappaP}{1.17}

\newcommand{\numKappaPN}{139}

\newcommand{\numGainRange}{0.03--10}

\newcommand{\numEtaCosRange}{0.998--0.999}

\newcommand{\numEtaPRange}{0.21--0.23}

\newcommand{\numEtaExitProduct}{0.52--0.55}

\newcommand{\numSoftmaxCos}{0.271}

\newcommand{\numLadderVerdict}{The collapse switches on at the rung that adds trainable hidden biases and is present at every rung after it; adding nonzero-mean targets alone, with zero-mean inputs and no biases, does not produce it}

\newcommand{\numCifarBaseAcc}{0.276}

\newcommand{\numCifarBPAcc}{0.343}

\newcommand{\numCifarPriorAcc}{0.349}

\newcommand{\numCifarCNNMinP}{0.009}

\newcommand{\numCifarCNNBPMinP}{0.756}

\newcommand{\numGaussRatioEarly}{1.00}

\newcommand{\numGaussRatioLate}{0.84}

\newcommand{\numGaussSharedHundred}{0.994}

\newcommand{\numGaussZeroMeanHundred}{0.0022}

\newcommand{\numGaussSharedLate}{0.515}

\newcommand{\numGaussZeroMeanLate}{0.401}

\newcommand{\numGaussRuns}{15}

\newcommand{\numAlignedKappa}{0.12}

\newcommand{\numResidBaseInit}{0.01}

\newcommand{\numCureCenterOverhead}{1.08}

\newcommand{\numCureCenterEOverhead}{0.95}

\newcommand{\numCurePriorOverhead}{0.92}

\newcommand{\numCureOutlrOverhead}{0.98}

\newcommand{\numCureBNOverhead}{1.10}

\newcommand{\numCureWhitenOverhead}{1.74}

\newcommand{\numCureMuonOverhead}{1.21}

\newcommand{\numFACos}{0.991}

\newcommand{\numDepthSixCos}{1.000}

\newcommand{\numDepthSixDuration}{520}

\newcommand{\numCNNConvOneCos}{0.951}

\newcommand{\numCNNConvOneBP}{0.634}

\newcommand{\numMSERatio}{0.36}

\newcommand{\numBCERatio}{1.28}

\newcommand{\numGenRCos}{0.85}

\newcommand{\numGenConditions}{18}

\newcommand{\numGenRCosCI}{0.30 to 0.96}

\newcommand{\numGenN}{102}

\newcommand{\numImbScratchCos}{0.938}

\newcommand{\numImbScratchBP}{0.386}

\newcommand{\numShiftCos}{0.03}

\newcommand{\numShiftEbarPost}{0.04}

\newcommand{\numShiftMinorityPre}{0.91}

\newcommand{\numShiftMinorityPost}{0.89}

\newcommand{\numShiftEbarAtShift}{0.11}

\newcommand{\numHeadMSECtrCos}{0.995}

\newcommand{\numHeadMSECtrEbar}{0.80}

\newcommand{\numHeadMSECtrWzeroCos}{0.378}

\newcommand{\numHeadMSECtrWzeroEbar}{0.02}

\newcommand{\numHeadMSECtrKappa}{1.02}

\newcommand{\numHeadSigmoidWzeroCos}{0.998}

\newcommand{\numAdamFastExit}{110}

\newcommand{\numAdamFastP}{0.046}

\newcommand{\numAdamSGDP}{0.214}

\newcommand{\numNoiseSlopeMin}{-0.503}

\newcommand{\numNoiseSlopeMax}{-0.493}

\newcommand{\numReviewConditionR}{0.91}

\newcommand{\numReviewSGDLearn}{537}

\newcommand{\numReviewAdamLearn}{117}

\newcommand{\numReviewAdamSlowLearn}{480}

\newcommand{\numReviewRawSoftmaxCos}{0.55}

\newcommand{\numReviewInitialDecode}{89.3}

\newcommand{\numReviewCollapseDecode}{88.6}

\newcommand{\numReviewFrozenRaw}{20.0}

\newcommand{\numReviewFrozenScaled}{80.1}

\newcommand{\numReviewReadoutMeanFifty}{0.94}

\newcommand{\numReviewWindowFixedR}{0.89}

\newcommand{\numReviewWindowRescaledR}{0.89}

\newcommand{\numReviewWindowFixedMAE}{0.105}

\newcommand{\numReviewWindowRescaledMAE}{0.107}

\newcommand{\numReviewRateHid}{-0.40}

\newcommand{\numReviewRateOut}{-0.60}

\newcommand{\numReviewRateRtwo}{0.98}

\newcommand{\numReviewRateRuns}{43}

\newcommand{\numReviewRateEligible}{72}

\newcommand{\numCostFiftyBase}{690}

\newcommand{\numCostFiftyBoth}{370}

\newcommand{\numSelfTermBase}{1.28}

\newcommand{\numSelfTermBP}{1.06}

\newcommand{\numSelfTermBPx}{0.07}

\newcommand{\numSelfTermAligned}{0.23}

\newcommand{\numBPxEbar}{1.50}

\newcommand{\numModelCondMAE}{0.10}

\newcommand{\numReadoutInitRaw}{63.4}

\newcommand{\numRMSFold}{5.9}

\newcommand{\numRMSRateFold}{35}

\newcommand{\numStdLOneCos}{0.80}

\newcommand{\numRawLOneCos}{0.90}

\newcommand{\numStdLOneP}{0.59}

\newcommand{\numRawLOneP}{0.71}

\newcommand{\numEbarHundred}{0.10}

\newcommand{\numEtilPlateauLo}{0.90}

\newcommand{\numEtilPlateauHi}{0.95}

\newcommand{\numSignCosEnd}{0.96}

\newcommand{\numSignPEnd}{0.05}

\newcommand{\numSignAcc}{0.75}

\newcommand{\numSignCenterCosEnd}{0.03}

\newcommand{\numSignCenterAcc}{0.90}

\newcommand{\numGainLowP}{0.98}

\newcommand{\numGainLowTime}{1897}

\newcommand{\numGainBaseTime}{537}

\newcommand{\numGainHighTime}{260}

\newcommand{\numGainHighP}{0.021}

\newcommand{\numCondDefaultAcc}{0.58}

\newcommand{\numCNNDeepP}{0.021}

\newcommand{\numCNNDeepBPP}{0.84}

\newcommand{\numAdamBPCos}{0.78}

\newcommand{\numAdamBPP}{0.70}

\newcommand{\numReadoutInitScaled}{82.9}

\newcommand{\numValidModelR}{0.99}

\newcommand{\numValidNullR}{0.67}

\newcommand{\numValidConds}{37}

\newcommand{\numSweepRMin}{0.98}

\newcommand{\numSweepRMax}{0.99}

\newcommand{\numSweepRClass}{0.10}

\newcommand{\numCenteredVarBase}{0.028}

\newcommand{\numCenteredVarBP}{0.83}

\newcommand{\numCenteredVarPrior}{0.97}

\newcommand{\numCenteredRankInit}{46}

\newcommand{\numCenteredRankBase}{4.3}

\newcommand{\numCenteredRankBP}{37}

\newcommand{\numCenteredRankCenterE}{4.6}

\newcommand{\numCenteredRankPrior}{5.3}

\newcommand{\numInstabLinearHzero}{17}

\newcommand{\numInstabLinearHend}{3.5\times10^{14}}

\newcommand{\numInstabLinearCenteredHmax}{0.6}

\newcommand{\numInstabReluHend}{3.6\times10^{6}}

\newcommand{\numInstabGeluHend}{3.3\times10^{7}}

\newcommand{\numLrDFAHighEighty}{130}

\newcommand{\numLrPriorHighEighty}{28}

\newcommand{\numLrBPHighEighty}{20}

\newcommand{\numLrSavingLow}{30}

\newcommand{\numLrSavingHigh}{78}

\newcommand{\numImbUncalCos}{0.96}

\newcommand{\numImbCalCos}{0.36}

\newcommand{\numImbUncalBalEarly}{0.30}

\newcommand{\numImbCalBalEarly}{0.21}

\newcommand{\numImbCalBalEnd}{0.89}

\newcommand{\numImbCalEbar}{0.03}

\newcommand{\numNokDFAZeroFastChance}{40}

\newcommand{\numNokDFAZeroSlowChance}{47}

\newcommand{\numNokBPFaninFastChance}{223}

\newcommand{\numNokBPFaninSlowChance}{10}

\makeatletter\@ifundefined{proposition}{\newtheorem{proposition}{Proposition}}{}\makeatother

\title{Common-Mode Collapse and Recovery\\in Direct Feedback Alignment}

\arxivversiontrue
\author{Varun Reddy$^{1}$, Bernardo L.\ Sabatini$^{1,2}$\thanks{Joint senior authors.}\quad\& Houman Safaai$^{1,*}$\thanks{Correspondence: \texttt{houman\_safaai@harvard.edu}}\\[0.5ex]
\normalfont\small $^{1}$Kempner Institute for the Study of Natural and Artificial Intelligence at Harvard University\\
\normalfont\small $^{2}$Department of Neurobiology, Howard Hughes Medical Institute, Harvard Medical School}

\iclrfinalcopy
\hypersetup{
  pdftitle={Common-Mode Collapse and Recovery in Direct Feedback Alignment},
  pdfauthor={Varun Reddy, Bernardo L. Sabatini, Houman Safaai}
}
\begin{document}
\maketitle
\pagestyle{plain}
% Shared manuscript body: edit section files to update both versions.
\begin{abstract}
Direct feedback alignment (DFA) trains hidden layers through fixed random projections of output error. With tanh hidden units and independent sigmoid outputs, plain stochastic gradient descent can stall near the loss of a constant predictor of class frequencies. We trace this stall to the error's common mode, the component shared across inputs. An exact mean--covariance decomposition separates a rank-one update formed by the mean teaching signal and mean presynaptic activity. Its leading component drives tanh units toward saturation. At initialization, random feedback provides no systematic correction of the shared error on average; readout learning limits its duration. A reduced model initialized from the network, without fitted parameters, predicts the concentration of activation sensitivity across \numModelConds\ settings. On MNIST, class decodability largely survives collapse, but readout learning remains slow at a fixed learning rate. Adam learns faster despite deeper collapse. Calibrating the baseline readout to the class prior suppresses collapse and speeds learning; weaker feedback trades less collapse for slower learning. Replacing errors by their signs sustains collapse; subtracting the signal's batch mean prevents sustained collapse and improves learning in the tested setting. Related effects occur in deeper and convolutional networks and on CIFAR-10, with severity and cost depending on the readout, optimizer and input statistics.

\end{abstract}
\section{Introduction}
\label{sec:intro}

How neural circuits assign errors to synapses is a central problem in theories of learning \citep{lillicrap2020}. Backpropagation (BP) uses downstream weights to compute hidden error signals \citep{rumelhart1986}. Direct feedback alignment (DFA) instead broadcasts output error through fixed random matrices \citep{nokland2016}. Like layerwise feedback alignment (FA), it avoids copying forward weights into the feedback pathway \citep{lillicrap2016}. Its performance varies across tasks and architectures \citep{bartunov2018,launay2020}. In particular, saturating tanh derivatives can weaken DFA gradients \citep{launay2019}, and plateaus or low-rank updates can arise \citep{refinetti2021,boeshertz2026}. We ask how a broadcast teaching signal can stall learning.

The original DFA experiments used independent sigmoid outputs with binary cross-entropy, and tanh hidden units to bound hidden updates that persist while the error is nonzero \citep{nokland2016}. With one-hot targets, initial predictions near one half leave a large mean error: a \emph{common mode} shared across inputs. Such an error arises whenever the mean prediction differs from the mean target, as can occur with imbalanced softmax or multi-label readouts. Fixed feedback sends it to each hidden unit with a different random coefficient. Together with mean presynaptic activity and trainable biases, this signal pushes tanh units toward saturation. We call the resulting suppression of activity differences across inputs \emph{representation collapse}. It occurs early and across classes, unlike the late within-class convergence studied as neural collapse \citep{papyan2020}; class information can nevertheless remain decodable. As the output layer, or \emph{readout}, learns the class frequencies (the \emph{class prior}), the mean drive weakens, but input-dependent learning can remain slow for hundreds of updates.

We derive a reduced model from the mean--covariance decomposition (Section~\ref{sec:theory}) and test its predictions across learning conditions (Section~\ref{sec:predictions}). Frozen-feature experiments distinguish decodability from learning speed (Section~\ref{sec:frozen}). We then assess the mechanism's scope and the costs of interventions across learning rules, architectures and readouts (Sections~\ref{sec:generality}--\ref{sec:cures}).

\section{Setting and the phenomenon}
\label{sec:setup}

\begin{figure*}[t]\centering\includegraphics[width=\textwidth]{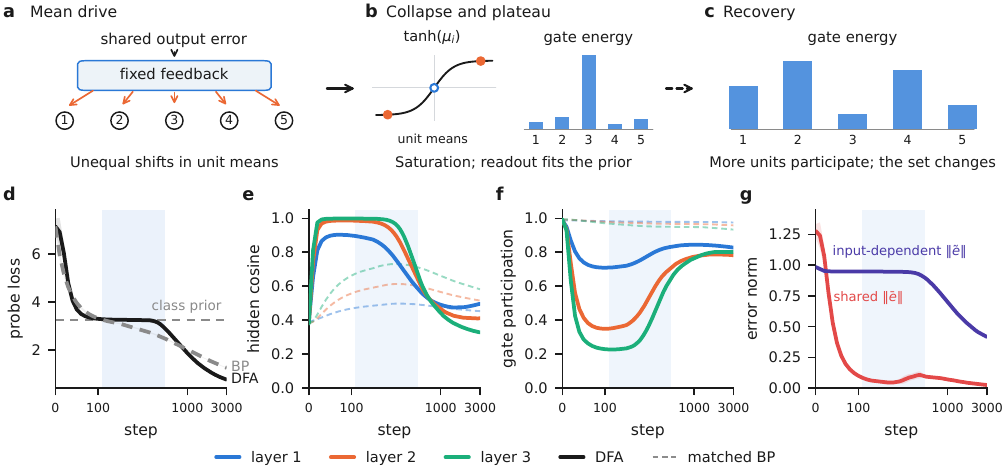}
\caption{\textbf{Collapse and recovery.} (a--c) Schematic: shared error shifts unit means toward tanh saturation (b: open, initial; filled, shifted), concentrating gate energy. More units participate during recovery and their identities change. Points and bars are illustrative; numbers identify units. The solid arrow represents the approximate collapse mechanism, the dashed arrow observed recovery beyond the model. (d) Probe loss for DFA and matched BP (same initialization and minibatches); the horizontal line is the class-prior loss and shading marks the mean detected plateau interval, repeated in e--g. (e) Mean pairwise hidden cosine and (f) gate participation per layer; colours denote layers, solid curves DFA and dashed curves BP. (g) Shared-error norm $\|\bar e\|$ and root-mean-square input-dependent error, both measured on the fixed probe. Time is linear through step 150 and logarithmic thereafter. Curves average \numHeadlineTraj\ DFA trajectories and five BP networks; bands in d and g show one standard deviation.}\label{fig:phenomenon}\end{figure*}

We first define the learning signals and activity measurements needed to describe collapse and recovery (Figure~\ref{fig:phenomenon}a--c).

\paragraph{Network and learning rules.}
In a feedforward network, layer $\ell$ has $d_\ell$ units, weights $W_\ell$, biases $b_\ell$, total inputs $a_\ell=W_\ell h_{\ell-1}+b_\ell$ (\emph{preactivations}) and activities $h_\ell=\phi(a_\ell)$. Here $\phi$ is the activation function, $h_0=x$, and $L$ is the number of hidden layers. The readout forms logits $z=W_{\rm out}h_L+b_{\rm out}$, the values before the output nonlinearity. For $C$ outputs, let $\hat y(x)\in\R^C$ denote the prediction and $y(x)\in\R^C$ the target. We use independent sigmoids with binary cross-entropy (BCE) summed over outputs, softmax with cross-entropy, or a linear readout with half squared error. Writing the per-example loss as $\mathcal L(x)$, the logit error is $e(x)=\partial\mathcal L(x)/\partial z=\hat y(x)-y(x)$. With stochastic gradient descent (SGD), DFA projects this error through a fixed matrix $B_\ell\in\R^{d_\ell\times C}$ and multiplies it by the activation derivative, or \emph{gate}. The resulting hidden teaching signal $\delta_\ell$ determines each local update:
\begin{equation}
\begin{aligned}
\delta_\ell(x)&=\phi'(a_\ell(x))\odot B_\ell e(x),\\
\Delta W_\ell&=-\eta\langle\delta_\ell h_{\ell-1}^{\top}\rangle,
\qquad \Delta b_\ell=-\eta\langle\delta_\ell\rangle.
\end{aligned}
\label{eq:dfa}
\end{equation}
Here $\Delta$ denotes a change in one update, $\odot$ denotes elementwise multiplication, $\langle\cdot\rangle$ is a minibatch mean, and $\eta=\eta_{\rm hid}$ is the hidden learning rate. Unless stated otherwise, entries of $B_\ell$ are independent $\mathcal N(0,g^2)$ draws, where $g$ is the feedback gain. The output layer receives its true gradient at learning rate $\eta_{\rm out}$. BP propagates error through the transposed forward weights; layerwise FA uses fixed random matrices between successive layers \citep{lillicrap2016}.

The baseline is MNIST \citep{lecun1998}, with pixels in $[0,1]$, a multilayer perceptron (MLP) with three tanh layers of width $d=300$, Xavier-uniform weights \citep{glorot2010}, zero biases, $g=1$, and SGD with $\eta=\eta_{\rm out}=10^{-3}$ and batch size $128$. The readout, loss, hidden nonlinearity and input scaling follow the original DFA experiments \citep{nokland2016} and direct random target projection (DRTP) \citep{frenkel2021}, which used adaptive optimizers and other initializations; in our reconstruction of N{\o}kland's protocol the DFA stall lasts only tens of updates, and at the larger tested rate BP also saturates (Appendix~\ref{app:nokland}). Diagnostics use a fixed $2{,}048$-example validation set, called the \emph{probe}. Appendix~\ref{app:protocol} gives the splits, architectures and protocol.

\paragraph{Measurements.}
Here $\E_x$ averages over probe inputs; $\|\cdot\|$ is the Euclidean norm and $\|\cdot\|_F$ the Frobenius norm of a matrix. We write $\ebar$ for mean error and $\etil(x)=e(x)-\ebar$ for its input-dependent part. We describe a representation by its mean pairwise cosine $\cos_\ell$ across inputs and its mean-activity energy $H_\ell=\Hl{\ell}$, with $\hb_\ell=\E_xh_\ell$. For unit $i$, let $\mu_{\ell i}=\E_x a_{\ell i}$ and $\sigma_{\ell i}^2=\operatorname{Var}_x(a_{\ell i})$; its \emph{gate energy} $u_{\ell i}=\E_x[\phi'(a_{\ell i})^2]$ is its mean squared sensitivity. For nonzero total gate energy, the gate participation ratio
\begin{equation}
 p_\ell=\frac{(\sum_i u_{\ell i})^2}{d_\ell\sum_i u_{\ell i}^2}
 \label{eq:participation}
\end{equation}
equals one when gate energies are equal and approaches $1/d_\ell$ when one unit dominates; equally weak gates also give high participation, so it does not measure feedback strength. We track activity--target covariance $\Lambda_\ell=\Cov(h_\ell,y)$, through its energy $\|\Lambda_\ell\|_F^2$. Covariances average centered outer products over the probe for diagnostics and over the minibatch for updates. To assess useful learning, we compare the DFA gradient estimate with the true BP weight gradient: their cosine $\cos\alpha_\ell$ measures direction, and projected descent $\Pi_\ell$ \citep{safaai2026}, the cosine times the DFA-to-BP norm ratio, predicts a first-order loss decrease from the hidden weight update when positive (Appendix~\ref{app:diagnostics}).

\paragraph{The baseline trajectory.}
DFA loss rapidly approaches that of a predictor returning the class frequencies and plateaus for several hundred steps before decreasing again; accuracy remains near chance during the plateau (Figure~\ref{fig:phenomenon}d). For sigmoid outputs, this prior loss is $\mathcal L_{\rm prior}=\sum_c\mathcal H(\pi_c)$, where $\pi=(\pi_c)$ is the vector of class frequencies and $\mathcal H(q)=-q\ln q-(1-q)\ln(1-q)$ is binary entropy. At this learning rate, matched BP passes through this loss without pausing but learns more slowly later. Along the DFA trajectory, raw DFA weight-gradient estimates are several times larger than BP gradients evaluated at the same network state (Figure~\ref{fig:gates}d).

During the plateau, top-layer hidden vectors become nearly parallel, unlike those of the matched BP network (Figure~\ref{fig:phenomenon}e). After subtracting each unit's mean activity, the minimum variance across inputs averages \numCenteredVarBase\ of its initial value, against \numCenteredVarBP\ for BP (Appendix~\ref{app:centered}). Gate participation falls in every hidden layer, most strongly in the deeper layers (panel~f), and the top-layer DFA gradient estimate is slightly anti-aligned with the BP gradient while the readout supplies the loss improvement (Figure~\ref{fig:useful_learning}c). As training recovers, cosine declines and participation rebounds. The shared error norm falls from \numEbarInit\ to \numEbarHundred\ within 100 updates, whereas the root-mean-square norm of input-dependent error stays between \numEtilPlateauLo\ and \numEtilPlateauHi\ throughout the plateau (panel~g). We now derive how the shared component drives collapse.

% Figure environments using explicit graphicx commands for arXiv.

\section{Theory: the common-mode term}
\label{sec:theory}

To explain collapse, we separate means from variation across inputs: $e(x)=\ebar+\etil(x)$ and $h_{\ell-1}(x)=\hb_{\ell-1}+\tilde h_{\ell-1}(x)$, where tildes denote deviations from the batch mean. Write the derivative gate as $\gamma_\ell(x)=\phi'(a_\ell(x))$, with mean $\bar\gamma_\ell$. This exact separation forms the basis of an approximate dynamical model.

\subsection{Exact decomposition and the common-mode closure}
\begin{proposition}[Mean and covariance parts of the DFA update]
\label{prop:rankone}
For any minibatch,
\begin{equation}
\begin{aligned}
\Delta W_\ell&=-\eta\Cov(\delta_\ell,h_{\ell-1})-\eta\bar\delta_\ell\hb_{\ell-1}^{\top},\\
\bar\delta_\ell&=\bar\gamma_\ell\odot B_\ell\ebar+r_\ell,
\qquad r_\ell=\big\langle(\gamma_\ell-\bar\gamma_\ell)\odot B_\ell\etil\big\rangle,
\end{aligned}
\label{eq:decomp}
\end{equation}
and $\Delta b_\ell=-\eta\bar\delta_\ell$. The second weight-update term has rank at most one. At initialization, conditional on $\ebar$, Gaussian feedback independent of the forward network gives independent entries $(B_\ell\ebar)_i\sim\mathcal N(0,g^2\|\ebar\|^2)$.
\end{proposition}

This is the classical mean--covariance decomposition of an outer product \citep{sejnowski1977}. The residual $r_\ell$ measures correlations between gate fluctuations and input-dependent error, which are weak at initialization in the baseline. A separate approximation treats the mean-error direction as nearly fixed during the early, approximately class-symmetric transient. Neglecting $r_\ell$ gives the \emph{common-mode closure}: the mean gated error is approximated by the product of mean gate and mean broadcast error. On the probe, $\|r_\ell\|$ is \numResidBaseInit\ of the leading term at initialization but exceeds it by step 300 (Appendix~\ref{app:residual}): the closure describes the onset of collapse, not the plateau.

The rank-one update shifts preactivation means and can also change their spread. Holding presynaptic activity fixed, we neglect spread changes, local covariance and the gate--error residual. With the tanh approximation $\bar\gamma_{\ell i}\approx\sech^2\mu_{\ell i}$, this gives
\begin{equation}
\frac{d\mu_{\ell i}}{dt}=-\eta\sech^2(\mu_{\ell i})(B_\ell\ebar)_i(H_{\ell-1}+1).
\label{eq:drift}
\end{equation}
Here $t$ counts updates, $\sech\mu=1/\cosh\mu$, and $+1$ comes from the bias update. Saturation slows fixed-direction drift to logarithmic growth. In baseline layer 3, upstream motion dominates the initial displacement; the mean term becomes dominant during early saturation (Appendix~\ref{app:drift_budget}). Exactly centering $h_{\ell-1}$ and freezing the postsynaptic bias $b_\ell$ removes local mean drift when presynaptic activity is fixed.

\subsection{Why the drive persists, and the collapse number}
Collapse also depends on how long the mean error acts. We combine the readout update with the local top-layer drift, using the same closure and omitting upstream motion and covariance terms. Let $\bar\Gamma_L=\operatorname{diag}(\bar\gamma_L)$ collect the mean derivative gates and $I$ be the $C\times C$ identity. The mean logits then obey
\begin{equation}
\frac{d\bar z}{dt}\approx-\Big[\eta_{\rm out}(H_L+1)I+\eta(H_{L-1}+1)\,W_{\rm out}\bar\Gamma_L^2B_L\Big]\ebar.
\label{eq:selfterm}
\end{equation}
For a positive multiple $B_L\propto W_{\rm out}^\top$, the hidden-layer matrix is positive semidefinite: within this approximation, hidden drift moves the mean logits in a direction that lowers the mean-only loss. This relation holds for BP and initially for pre-aligned DFA. Random feedback has no systematic restoring effect at initialization when averaged over feedback draws; the readout supplies the systematic restoring term in this approximation. The shared error falls more quickly under pre-aligned feedback and BP with an amplified readout, consistent with this distinction (Appendix~\ref{app:alignment_controls}).

Keeping only the readout term, balanced classes with $C>2$, zero initial logits and constant $H_L$ give an effective slope for the sigmoid $\sigma(z)=(1+e^{-z})^{-1}$, $\sigma'_{\rm eff}=(1/2-1/C)/\ln(C-1)$ and initial error norm $\|\ebar_0\|=\sqrt C(1/2-1/C)$ (Appendix~\ref{app:dose}). Together with the drift, this gives the initialization estimate
\begin{equation}
\hat\kap_\ell=
\frac{g\|\ebar_0\|(H_{\ell-1}(0)+1)}{\sigma'_{\rm eff}(H_L(0)+1)(\eta_{\rm out}/\eta)}.
\label{eq:kappa}
\end{equation}
This dimensionless \emph{collapse number} is unchanged by scaling both learning rates; at fixed effective output slope, gain and initial mean error enter through their product. It ignores the evolution of activity energies and input-dependent learning. We therefore also use the reduced-model prediction below and the measured \emph{delivered dose} along the initial mean-error direction,
\begin{equation}
 \kap_\ell(t)=g\sum_{s<t}\eta\langle\ebar_s,\hat e_0\rangle(H_{\ell-1}(s)+1),
 \qquad \hat e_0=\ebar_0/\|\ebar_0\|,
 \label{eq:delivered}
\end{equation}
where $s$ indexes updates, $\hat e_0$ is the unit initial-error vector, and $\langle u,v\rangle=u^\top v$ denotes a vector inner product, distinct from the minibatch average $\langle\cdot\rangle$. For comparisons of initial collapse, the \emph{transient dose} stops this sum when the probe mean-error norm falls to $15\%$ of its initial value (Appendix~\ref{app:dose}). % provenance-ok: protocol constant (15 percent cutoff)

In deeper layers, saturation of presynaptic activity also narrows the within-unit preactivation distributions. For small spread, large width and negligible initial means, averaging over unit means gives $p_\ell\sim(15/8)\sqrt{2/\pi}/\kap_\ell$ at large dose, bounded below by $1/d_\ell$ (Appendix~\ref{app:quadrature}). Gaussian reconstructions from measured means and spreads reproduce early participation (Appendix~\ref{app:gate_moments}).

\subsection{Coupled dynamics across layers}
The initialization estimate reverses the measured ordering of dose across layers because activity energies evolve: as a layer saturates, $H_\ell$ grows toward its width, increasing the drive on the next layer and accelerating the readout's learning of the prior.

\begin{figure*}[t]\centering\includegraphics[width=\textwidth]{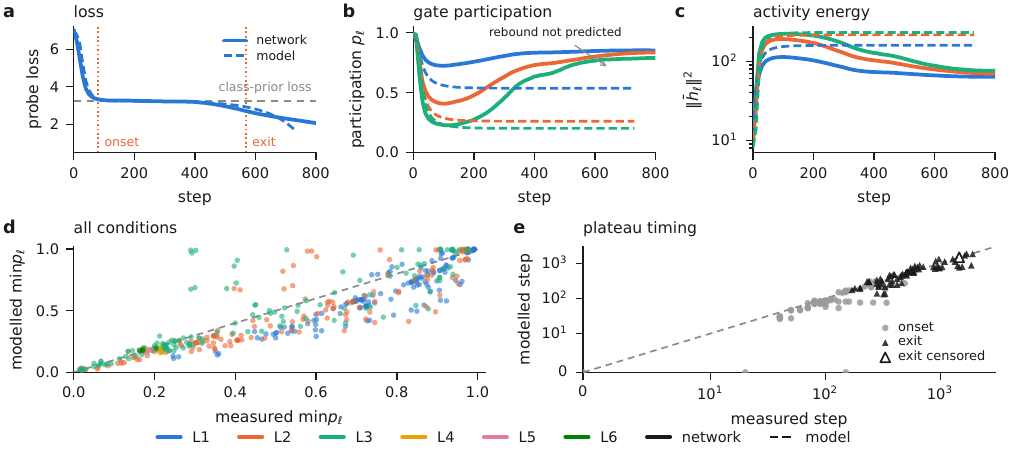}
\caption{\textbf{The model predicts early concentration and approximate exit times, but not participation recovery.} (a--c) Model (dashed) and matched network (solid), both using initialization/feedback seeds $(0,0)$: loss, gate participation and mean-activity energy. Dotted vertical lines in a mark predicted onset and exit, the exit from the covariance extension of Section~\ref{sec:escape}. (d) Participation minima within the first 300 updates across \numModelConds\ settings and three seed pairs; colours denote layers (L1--L6). Points with model values near one but low measured minima are two-class and squared-error settings, for which the model predicts almost no drive. (e) Paired onset and exit times; the open triangle gives a lower bound for an exit not reached within the model horizon. Diagonals in d,e indicate equality.}\label{fig:reduced}\end{figure*}

Gaussian quadrature computes mean activities and gates by averaging over each unit's evolving Gaussian preactivation distribution. The reduced model uses a shared within-unit variance $\sigma_\ell^2$, fixed in layer 1 and propagated as $\sigma_\ell^2=q_\ell\bar u_{\ell-1}\sigma_{\ell-1}^2$ for $\ell\ge2$. Here $q_\ell=d_{\ell-1}\operatorname{Var}(W_{\ell,ij})$ is fixed at initialization and $\bar u_\ell=d_\ell^{-1}\sum_i u_{\ell i}$ is mean gate energy. Retaining only the readout term of Equation~\eqref{eq:selfterm} closes the system, initialized from the network without fitted parameters (Appendix~\ref{app:model}). It captures the baseline plateau loss, onset and layer ordering of participation minima (Figure~\ref{fig:reduced}a--d). The covariance extension below adds exit predictions.

\subsection{Recovery and the deeper-but-shorter effect}
\label{sec:escape}
Ending the mean drive does not immediately restore learning: the network remains close to the prior predictor. Near that predictor, the input-dependent readout gradient is approximately $-\Lambda_L^\top$, so its first-order contribution to loss decrease is $\eta_{\rm out}\|\Lambda_L\|_F^2$. This energy falls sharply while a scale-normalized statistic changes little (Table~\ref{tab:E7}; Section~\ref{sec:frozen}). During recovery, label covariance grows again, beginning in earlier layers. A phenomenological covariance model initialized from $\Cov(x,y)$ predicts when this growth lowers loss (Figure~\ref{fig:reduced}e; Appendix~\ref{app:escape}).

Coupled changes in features and readout make the plateau duration $T_{\rm plateau}$, from onset to exit (Appendix~\ref{app:protocol}), depend on both rates. Under plain SGD, gain and hidden rate enter only through $a=\eta_{\rm hid}g$, and a fit to the detected plateaus gives
\begin{equation}
 T_{\rm plateau}\propto a^{\numReviewRateHid}\,\eta_{\rm out}^{\numReviewRateOut}
 \label{eq:recovery}
\end{equation}
(Table~\ref{tab:review_rate_fit}). The exponents sum to approximately $-1$, as the measured $1/\eta$ scaling requires when both rates change together. Over this tested range, increasing gain deepens the initial collapse (Section~\ref{sec:predictions}) while shortening the plateau. Recovery also changes the gates: participation rebounds, the participating units turn over, and the within-unit spread grows; the reduced model does not reproduce these features (Appendix~\ref{app:gates}).

\section{Predictions and their tests}
\label{sec:predictions}

The account predicts that the strength and duration of the shared drive affect collapse differently. We test this by varying feedback gain, learning rates, initial predictions and input means (Appendix~\ref{app:results}).

\paragraph{Reduced-model predictions.}
Across \numModelConds\ distinct settings, the condition-averaged correlation between predicted and measured participation minima is $r=\numReviewConditionR$ (mean absolute error \numModelCondMAE). The model outperforms the initialization estimate $\min(1,1.5/\hat\kap_\ell)$ obtained from Equation~\eqref{eq:kappa} and the large-dose asymptote (Table~\ref{tab:submission_null}). Its errors reveal the approximation's limits: it overestimates shallow-layer collapse in the baseline (Figure~\ref{fig:reduced}b) and misses concentration in two-class and squared-error settings (panel~d). Table~\ref{tab:model_coverage} retains detection failures and censored exits. The recovery extension predicts broad variation in exit times (panel~e), including on untested width--depth combinations; removing local covariance growth predicts no exits in those new settings (Appendix~\ref{app:recovery_tests}). % provenance-ok: closed-form constant 1.5 of the large-dose asymptote

\begin{figure*}[t]\centering\includegraphics[width=\textwidth]{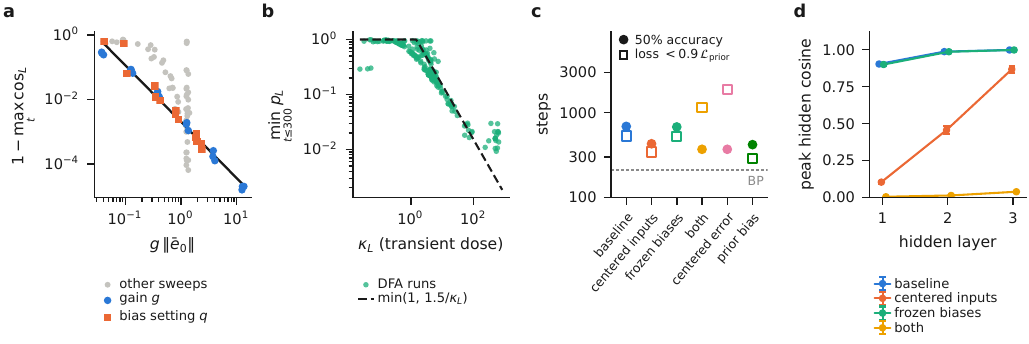}
\caption{\textbf{Scaling, learning cost and mean-channel controls.} (a) $1-\max_t\cos_L$ against $g\|\ebar_0\|$: smaller values indicate higher cosine. Gain and initial-prediction sweeps are coloured, other sweeps grey; the line fits the coloured points, counting their shared baseline once. (b) Deepest-layer participation minimum within 300 updates against measured transient dose; dashed line: large-dose prediction capped at one. (c) Steps to 50\% probe accuracy (filled; median) and to a sustained loss below $0.9\mathcal L_{\rm prior}$ (open; mean) for the controls in d, centered output error and prior-bias initialization; dotted line: BP at 50\% accuracy. (d) Peak cosine in each layer over the first 800 updates: input centering and frozen hidden biases jointly prevent collapse. Error bars in d: one standard deviation; trajectories in Figure~\ref{fig:dissection}a--d.}\label{fig:master}\end{figure*}

\paragraph{Learning rate, gain and initial predictions.}
Changing both learning rates together over a tenfold range leaves peak hidden cosine and participation minima within seed noise, while exit time scales as $1/\eta$ (Appendix~\ref{app:scaling}): the rates change the trajectory's timescale with little effect on collapse severity. Varying gain and the nominal initial sigmoid probability $q$, set by output bias $\ln[q/(1-q)]$, changes the drive; for constant initial prediction, $\|\ebar_0\|=\sqrt C|q-1/C|$. Both sweeps follow one power-law trend in $g\|\ebar_0\|$: stronger drive produces more nearly parallel hidden activities (Figure~\ref{fig:master}a). This scaling is empirical because $q$ also changes the readout timescale (Appendix~\ref{app:dose}), and other sweeps at similar $g\|\ebar_0\|$ scatter widely. At large delivered dose, participation decreases on the scale predicted by Gaussian quadrature (panel~b).

 % provenance-ok: learning-time criteria

\paragraph{Readout and input means.}
Balanced softmax has zero mean error at uniform prediction, and with standardized inputs its collapse is weak (peak cosine \numSoftmaxCos, or \numReviewRawSoftmaxCos\ with raw pixels; Table~\ref{tab:review_heads}); drawing from class 0 with probability $0.7$ and from the original distribution otherwise restores it (\numImbScratchCos, against \numImbScratchBP\ for BP; Table~\ref{tab:E1}). Per-pixel centering sets $\bar x$ to zero on the training split. It prevents layer-1 collapse but only reduces deeper collapse. Freezing hidden biases alone has little effect; combining both prevents collapse throughout the network (Figure~\ref{fig:master}d). Global standardization raises the input mean energy and deepens layer-1 gate concentration despite lowering peak cosine (Appendix~\ref{app:alignment_controls}). % provenance-ok: class-0 mixture weight

\paragraph{Learning cost.}
Collapse delays early discrimination (Figure~\ref{fig:master}c; Table~\ref{tab:submission_cost}). Centering inputs with frozen biases, or centering output error, prevents collapse and reaches 50\% accuracy after \numCostFiftyBoth\ rather than \numCostFiftyBase\ updates. However, unsaturated layers keep $H_L$ small (\numCenterEHL\ versus \numBaseHLThree\ at step 300 under error centering), which slows the readout's fit to the prior, so the time to 80\% accuracy improves little. Prior-bias initialization suppresses the drive while speeding both early discrimination and loss reduction, with unchanged final accuracy. % provenance-ok: accuracy thresholds

For comparisons across conditions, we define \emph{learning time} as the first probe loss below $0.9\mathcal L_{\rm prior}$ that remains below it for 50 updates. Weaker feedback prevents collapse but increases this learning time: at $g=0.03$, close to the feedback scale of \citet{nokland2016} at this width, minimum participation is \numGainLowP\ but learning time is \numGainLowTime\ rather than \numGainBaseTime\ updates (Appendix~\ref{app:scaling}). Thus reduced collapse need not mean faster learning: saturation increases mean-activity energy and can accelerate the fit to the prior while weakening input sensitivity. The delay persists at larger learning rates: at $\eta=0.1$, the largest rate tested, no run diverges, and plain DFA needs \numLrDFAHighEighty\ updates to reach 80\% accuracy, against \numLrPriorHighEighty\ with a calibrated readout and \numLrBPHighEighty\ for BP (Appendix~\ref{app:scaling}). % provenance-ok: accuracy and sustained-loss thresholds and gain values

\paragraph{Unit drift and feedback orientation.}
A nearly class-symmetric error predicts drift along $B_\ell\mathbf1$, with $\mathbf1$ the all-ones vector; measured unit mean shifts are anticorrelated with it in all three layers (Table~\ref{tab:review_unit_drift}), as Equation~\eqref{eq:drift} predicts. Pre-aligned feedback at the Frobenius norm of the random draw delivers a dose of only \numAlignedKappa\ rather than \numKappaThree\ and, like BP with an amplified readout, learns quickly with little collapse (Figure~\ref{fig:dissection}f,g), consistent with Equation~\eqref{eq:selfterm}. Reorienting feedback also changes $B_\ell\ebar$, so this control does not isolate the restoring contribution.

\begin{figure*}[t]\centering\includegraphics[width=\textwidth]{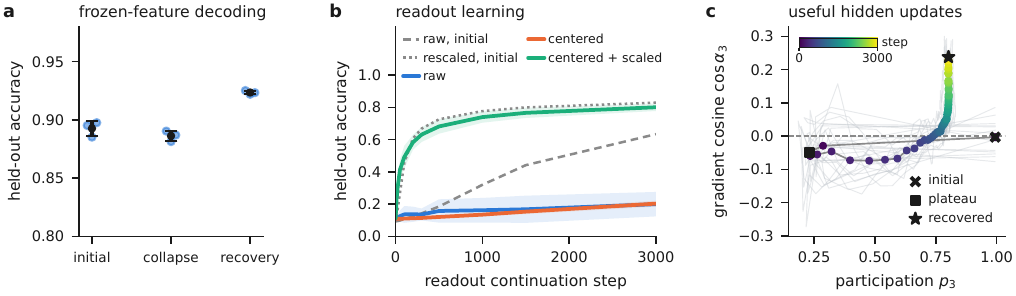}
\caption{\textbf{Decodability, readout learning and useful descent.} (a) Least-squares ridge decoding of one-hot targets on separate evaluation examples at three frozen layer-3 checkpoints. (b) Sigmoid/BCE readout continuation from the collapse checkpoint at nominal SGD rate $10^{-3}$ with raw, centered and centered/rescaled features; transformed parameters preserve the initial predictions. Grey curves continue the same raw (dashed) and centered, rescaled (dotted) readouts from the initial checkpoint. Curves and error bars in a,b show means and one standard deviation over three seed pairs. (c) Baseline gradient cosine against participation across 15 trajectories: grey curves are individual runs, coloured points follow the mean in time order; markers denote initialization, step 100 and step 3000.}\label{fig:useful_learning}\end{figure*} % provenance-ok: continuation rate and baseline cohort

\section{Decodable features can support slow learning}
\label{sec:frozen}

The covariance measurements suggest that collapse weakens learning signals while preserving class structure. We freeze the baseline network at initialization, at its early participation minimum, and after recovery. A least-squares ridge readout fitted to one-hot targets retains \numReviewCollapseDecode\% held-out accuracy at collapse, close to \numReviewInitialDecode\% at initialization (Figure~\ref{fig:useful_learning}a).

Learning from the collapsed features is nevertheless slow. With identical minibatches and a fixed nominal SGD rate, a sigmoid readout continued on them reaches \numReviewFrozenRaw\% held-out accuracy after 3,000 updates, against \numReadoutInitRaw\% from the initial features (panel~b). Collapsed features are \numRMSFold-fold smaller in root-mean-square (RMS) variation, which lowers the readout's effective learning rate about \numRMSRateFold-fold. Centering them changes little, whereas dividing centered features by their RMS and transforming the readout parameters to preserve its initial predictions raises accuracy to \numReviewFrozenScaled\%, close to \numReadoutInitScaled\% for rescaled initial features. This comparison identifies an amplitude bottleneck at a fixed optimizer scale despite retained class decodability (Appendix~\ref{app:frozen}). Adam \citep{kingma2015} normalizes updates coordinatewise and learns faster despite deeper gate concentration. Across three matched seed pairs, its learning time is \numReviewAdamLearn\ rather than \numReviewSGDLearn\ updates, with minimum participation \numAdamFastP\ rather than \numAdamSGDP\ (Appendix~\ref{app:optimizer}).

Recovery also involves a change in the direction of hidden-layer updates. Participation recovers while the top-layer gradient cosine is still near zero, and alignment then rises at nearly fixed participation (panel~c). Participation therefore does not determine update direction; describing recovery requires tracking task-dependent covariance and alignment alongside unit sensitivity.

\section{Generality and limits}
\label{sec:generality}

We next ask whether the transient depends on the baseline architecture and task, and what changes when the shared signal persists.

\paragraph{Rules, architectures and objectives.}
The same signature appears under layerwise FA (peak top-layer cosine \numFACos), in a six-layer MLP whose layers 4--6 reach cosine \numDepthSixCos, and in a small convolutional network (CNN), whose deepest block reaches minimum participation \numCNNDeepP\ against \numCNNDeepBPP\ for one matched BP run (Figure~\ref{fig:generality}; Appendix~\ref{app:generality}). Changing the objective changes the transient: an augmented twelve-output multi-label target also produces a prior-loss plateau, whereas large-target squared error produces only a brief interruption. With rectified linear units (ReLU), mean drift instead leaves many units inactive or always active, more often than in the matched BP control, and across settings the early ratio of mean to input-dependent error correlates with peak cosine (Appendix~\ref{app:generality}). % provenance-ok: six-layer architecture

\paragraph{CIFAR-10 and the size of the input mean.}
On CIFAR-10 \citep{krizhevsky2009}, a tanh CNN with a sigmoid readout reaches minimum participation \numCifarCNNMinP\ under DFA against \numCifarCNNBPMinP\ under BP (Table~\ref{tab:cifar_controls}); because BP also has high cosine on uncentered images (Figure~\ref{fig:cifar}a), participation is the more informative diagnostic here. With standardized inputs, balanced softmax largely avoids collapse, while imbalanced sampling restores it (panels~b,c). The larger CIFAR-10 input mean also drives collapse in MLP layer 1 (panel~d). Calibrating initial logits toward the class prior reduces mean error without preventing collapse, although prior-bias initialization raises CNN accuracy at step 3000 from \numCifarBaseAcc\ to \numCifarPriorAcc\ (BP: \numCifarBPAcc). Combining error and input centering reduces collapse in both networks. CIFAR-100 shows high transient cosine even under softmax with small initial error (Appendix~\ref{app:batch_test}). These controls show why output calibration alone can be insufficient when input means are large.

\paragraph{A common mode that does not decay.}
For sigmoid outputs and binary targets, sign-error feedback broadcasts $\operatorname{sign}(e)=\mathbf 1-2y$. Before gating, this is a fixed target projection, as in DRTP \citep{frenkel2021}, plus a constant. Its population mean $\mathbf1-2\pi$ does not shrink as the readout learns. With this signal, the MNIST baseline stays collapsed through 3,000 updates (final cosine \numSignCosEnd, participation \numSignPEnd, accuracy \numSignAcc), prior-bias initialization leaves its hidden updates unchanged, and subtracting its batch mean prevents sustained collapse and improves learning (cosine \numSignCenterCosEnd, accuracy \numSignCenterAcc; Appendix~\ref{app:sign}). % provenance-ok: training horizon

\section{Interventions and their costs}
\label{sec:cures}

The mechanism suggests three interventions: subtract the mean error, center presynaptic activity, or shorten the error's duration; Figure~\ref{fig:cures}a--c compares them in three networks.

\paragraph{Centering the broadcast error.}
Subtracting $\langle e\rangle$ before projection removes the mean-error contributions to the rank-one update and gate covariance (Equation~\eqref{eq:shared-error-update}). The gate--error residual can still contribute to the mean teaching signal. Error centering prevents strong convergence of hidden activities in the tanh MNIST baseline (Figure~\ref{fig:dissection}e); centering the teaching signal after the derivative gate is less effective. A faster readout offsets the slower prior fit under error centering (Section~\ref{sec:predictions}; Appendix~\ref{app:interventions}). With ReLU, Gaussian error linear units (GELU) \citep{hendrycks2016}, or linear units, however, error centering can drive loss far above the constant-predictor baseline as mean activity grows (Appendix~\ref{app:intervention_comparisons}). Input centering suppresses this extreme loss growth, although the linear network's BCE remains close to the constant-predictor loss (Table~\ref{tab:E10}).

\paragraph{Calibration and activity centering.}
Initializing output biases near $\operatorname{logit}(\pi_c)=\ln[\pi_c/(1-\pi_c)]$, the prior initialization used for class-imbalanced detection \citep{lin2017}, suppresses MNIST collapse and shortens learning; a faster readout shortens the drive. Neither suffices in the CIFAR-10 CNN. On the activity side, per-pixel input centering \citep{lecun1998efficient} protects the first layer, and batch normalization \citep{ioffe2015} of hidden preactivations reduces collapse; since normalization also rescales and shifts, this does not isolate mean subtraction.

\paragraph{Changing update geometry.}
Muon-style orthogonalization of hidden updates \citep{muon2024}, which \citet{boeshertz2026} applied to FA, reduces collapse. An adaptation of the error-side conditioner of \citet{safaai2026}, which suppresses dominant teaching-signal directions through an inverse second moment, retains high cosine at every tested ridge regularization strength and, at the default strength, lowers step-3000 accuracy to \numCondDefaultAcc\ (Figure~\ref{fig:cures}d; Appendix~\ref{app:interventions}).

\section{Related work}
\label{sec:related}

Our account concerns the cause and duration of early saturation, a long-recognized source of plateaus in sigmoid networks \citep{lee1993,vitela1997}. Under BP, sigmoid hidden units can saturate while the readout learns its biases, then desaturate slowly \citep{glorot2010}. Saturating derivatives also weaken tanh DFA gradients \citep{launay2019} and can halt learning under DRTP \citep{frenkel2021}. We identify a mean-driven contribution under random feedback, predict the concentration of sensitivity and approximate exit times, and test interventions on it.

This early transient complements accounts of alignment. \citet{refinetti2021} establish align-then-memorise dynamics; our stall precedes substantial alignment (Appendix~\ref{app:demarcation}). \citet{boeshertz2026} find low-rank error and update dynamics in layerwise FA and improve learning through orthogonalization and activity normalization; whether a common mode contributes there is open. \citet{lee2026} restore weak weight and gradient alignment when fine-tuning with DFA by constructing feedback from pretrained weights and re-initializing forward weights to match it. Our aligned control instead starts from an untrained network. \citet{safaai2026} condition the activity and error factors of the DFA update through their second moments; we study their first moments before alignment.

Centering has several antecedents. Nonzero-mean inputs bias gradient updates \citep{lecun1998efficient}, and \citet{schraudolph1998} centers activities, errors and activation slopes. Concurrently, \citet{yamada2026} subtract the batch mean of one-vs-all errors in a dual-stream network with structured routing, noting its strong class-wise offset early in training; we show how such an offset saturates units under random feedback. The mean--covariance separation is classical \citep{sejnowski1977,sejnowski1989,dayan2001}, and centered modulatory signals matter in reward-dependent plasticity \citep{fremaux2010,gerstner2018}. The distinction between sensitivity and useful learning connects to dormant units and plasticity loss \citep{sokar2023,lyle2024}.

\section{Discussion}
\label{sec:discussion}

The early DFA stall reflects competition between shared drive and input-dependent learning. With random feedback, mean output error can push units toward saturation before the readout learns the class prior. Our local approximation explains why this drift initially lacks the systematic restoring contribution supplied by transposed feedback. Class information remains decodable, but reduced feature variation weakens readout gradients. Learning resumes as covariance grows, a changing set of units regains sensitivity, and updates align with the loss gradient.

The controls suggest ways to limit this transient. Prior-calibrated initialization speeds learning in the MNIST baseline at no extra training cost, but is insufficient on CIFAR-10. Input centering protects early layers and prevents the observed instabilities under error centering with non-saturating units. For the tested sign-error rule, calibration cannot reduce the mean broadcast signal; centering it prevents sustained collapse and improves learning.

Our evidence comes mainly from small networks trained with plain SGD at fixed learning rates. Adaptive optimizers can shorten the transient (Appendices~\ref{app:optimizer} and~\ref{app:nokland}). The reduced model captures early collapse but not participation rebound; its recovery extension is phenomenological. These findings support a mechanism for early dynamics, rather than a general account of DFA performance. Whether biological circuits regulate a similar interaction between shared teaching signals and input-dependent plasticity remains open.

% Start submission statements on a new page in the anonymous version.
% The author-visible preprint keeps them with the preceding discussion.
\ificlrfinal\else\clearpage\fi
\subsection*{Reproducibility statement}
The supplementary code contains the training configurations, reduced model, tests and analysis scripts. Appendix~\ref{app:protocol} specifies the data splits and measurements; Appendix~\ref{app:derivations} states the model assumptions and derivations. Figures and numerical tables are generated from the experiment records.
\ifarxivversion
Code and reproduction materials will be made available at \href{https://github.com/KempnerInstitute/DFA-Stall}{\texttt{github.com/KempnerInstitute/DFA-Stall}}.
\fi

% Retain this disclosure in ICLR review and camera-ready builds.
\ifarxivversion\else
\subsection*{AI use statement}
Generative AI assisted coding, experiment execution, figure preparation, literature search and manuscript editing. The authors reviewed the outputs, checked numerical results and references, and take responsibility for the manuscript and accompanying artifacts.
\fi

% Funding and computing acknowledgement matches the nDFA paper; author-visible builds only.
\ificlrfinal
\subsection*{Acknowledgements}
This work has been made possible in part by a gift from the Chan Zuckerberg Initiative Foundation to establish the Kempner Institute for the Study of Natural and Artificial Intelligence at Harvard University. The computations in this paper were run on the Kempner Institute AI cluster at Harvard University.
\fi

\bibliography{refs}
\bibliographystyle{iclr2027_conference}
\newpage
\appendix
\setcounter{topnumber}{2}
\setcounter{dbltopnumber}{2}
% Keep float-only pages together at the top instead of stretching their gaps.
\makeatletter
\setlength{\@fptop}{0pt}
\setlength{\@fpsep}{20pt}
\makeatother
\section*{Supplementary material}
The supplement follows the main paper's argument. Appendix~\ref{app:protocol} defines the experimental protocol and measurements; Appendix~\ref{app:derivations} gives the mathematical model. Appendix~\ref{app:results} tests the mechanism, approximations and predictions. Appendix~\ref{app:representation} examines decodability and gate recovery, and Appendix~\ref{app:controls} establishes the scope and costs of the interventions. Numerical comparisons appear beside the relevant analyses. Complete condition summaries, including conditions not tabulated here, are supplied with the code.
\section{Experimental protocol and reproducibility}
\label{app:protocol}

This section specifies the data and network choices, defines the diagnostics used to separate activity changes from useful learning, and gives the intervention implementations and computational costs.

\subsection{Data, networks and randomness}
We use MNIST \citep{lecun1998}, Fashion-MNIST \citep{xiao2017fashion}, CIFAR-10 and CIFAR-100 \citep{krizhevsky2009} through \texttt{torchvision}. After any class restriction, the last $10{,}000$ examples of the official training set form the validation split; the remainder is used for training. The official test sets are not used. The diagnostic probe contains $2{,}048$ validation examples sampled with replacement according to the training sampler. Its indices remain fixed within a run, including after a label-prior shift; the probe then retains the pre-shift sampling distribution. Consequently, reported accuracy is a validation diagnostic, not a test-set benchmark. % provenance-ok: split and probe protocol

Default preprocessing divides pixels by 255. Global standardization subtracts one scalar training mean and divides by one scalar standard deviation. Per-pixel centering subtracts the training mean image; per-pixel standardization also divides by the pixelwise standard deviation plus $10^{-3}$. Only per-pixel centering makes the training mean vector zero, up to numerical precision; validation and minibatch means can remain nonzero. In imbalanced experiments, mixture weight $p_0$ selects a class-0 example, and the remaining weight selects from the original data distribution. Thus $p_0=0.7$ is a mixture parameter, not the final class-0 frequency. % provenance-ok: preprocessing and mixture protocol
The headline softmax conditions use global standardization; the batch-size softmax sweep retains raw pixels. Table~\ref{tab:review_heads} compares both with matched horizons and seeds. Restricting classes before the fixed validation split also changes the training-set size. The class-count sweep therefore checks the initialization formula without isolating class count as a cause of later learning speed. Imbalanced results are accompanied by the majority-class reference and balanced accuracy in Table~\ref{tab:review_imbalance}.

The base MLP has three hidden layers of width 300; depth and width sweeps change these values as specified in the configuration files. The small CNN has two $3\times3$ convolutions with 16 and 32 channels, padding one and strides one and two, followed by a 128-unit hidden fully connected layer and the output layer. Each hidden block uses the selected activation. DFA broadcasts to every flattened block activation; the local convolutional gradient then aggregates spatial contributions through weight sharing. The fully connected derivations describe the MLP, while CNN results test the diagnostic signature empirically. % provenance-ok: architecture protocol

For squared error, targets are scaled one-hot class vectors, optionally centered across classes. The multi-label readout retains ten class indicators and adds two binary targets, each active for three randomly selected classes. Accuracy always uses the original class outputs. Initialization and minibatch order share an initialization seed; feedback uses an independent seed. Main DFA conditions use five initializations and three feedback draws each (15 trajectories); BP uses the five corresponding initializations. Sweeps generally use three initialization--feedback pairs, with the exact count reported in each table. % provenance-ok: target construction and seed protocol

\paragraph{Readout functions and losses.}
For sigmoid outputs, $\hat y_c=\sigma(z_c)$ with $\sigma(z)=(1+e^{-z})^{-1}$; their probabilities are independent and need not sum to one. Softmax instead gives $\hat y_c=\exp(z_c)/\sum_j\exp(z_j)$. Their per-example losses are $-\sum_c[y_c\log\hat y_c+(1-y_c)\log(1-\hat y_c)]$ and $-\sum_c y_c\log\hat y_c$, respectively. For linear outputs, $\hat y=z$ and the loss is $\|z-y\|^2/2$. Each gives the error $e=\hat y-y$ used in Equation~\eqref{eq:dfa}; training averages the loss over a minibatch.

\subsection{Diagnostics and notation}
\label{app:diagnostics}
Unless a configuration overrides the interval, we log the probe loss, accuracy, mean error and layer statistics every ten updates, and the more expensive diagnostics every fifty. Per-unit means, spreads and gate energies are saved at fixed snapshots. Tables report means $\pm$ one standard deviation across runs unless stated otherwise. Norms of input-dependent error mean the root mean square, $(\E_x\|\etil(x)\|^2)^{1/2}$. % provenance-ok: logging protocol

\paragraph{Averages and symbols.}
Bars and $\langle\cdot\rangle$ in an update refer to its training minibatch. Diagnostic expectations $\E_x$ instead use the fixed probe; expectations over random feedback or sampled batches are stated separately. Layer indices run from $1$ to $L$, unit indices from $1$ to $d_\ell$, and subscript $0$ denotes initialization when attached to an error or time-dependent statistic. The notation below distinguishes measured unit moments from the approximation used in the reduced model.
\begingroup\small
\begin{longtable}{>{\raggedright\arraybackslash}p{.28\linewidth}>{\raggedright\arraybackslash}p{.65\linewidth}}
\caption{Notation used in the analysis. A bar denotes an average over inputs, except for $\bar u_\ell$ and $\overline{u^2}$, which average gate energies across units.}\label{tab:notation}\\
\toprule Symbol & Meaning\\\midrule
$C,\;L,\;d_\ell$ & Number of outputs, hidden layers and units in layer $\ell$; $i$ indexes units.\\
$x,\;h_\ell,\;a_\ell$ & Input, layer activity vector and preactivation vector; $h_0=x$.\\
$z,\;\pi,\;\mathcal L_{\rm prior}$ & Readout logits, class-frequency vector and loss of the constant prior predictor.\\
$\hat y,\;y,\;e$ & Prediction, target and output error $e=\hat y-y$.\\
$\bar e,\;\tilde e,\;\hat e_0$ & Mean error, deviation $e-\bar e$ and unit vector along the initial mean error.\\
$W_\ell,\;b_\ell,\;B_\ell$ & Forward weights, biases and fixed feedback matrix.\\
$\gamma_\ell,\;\delta_\ell$ & Activation derivative and gated teaching signal $\gamma_\ell\odot B_\ell e$.\\
$\bar\Gamma_\ell,\;r_\ell$ & Diagonal matrix of mean gates and the gate--error covariance residual in Equation~\eqref{eq:decomp}.\\
$\mu_{\ell i},\;\sigma_{\ell i}$ & Mean and standard deviation of one unit's preactivation across inputs.\\
$\sigma_\ell^2$ & Shared within-unit variance assumed by the reduced model; distinct from the measured $\sigma_{\ell i}^2$.\\
$\sigma(z),\;\sigma'_{\rm eff}$ & Sigmoid output function and its effective slope in the fixed-activity dose estimate; distinct from preactivation spread.\\
$H_\ell$ & Squared norm of the activity mean, $\|\bar h_\ell\|^2$; $H_0=\|\bar x\|^2$.\\
$u_{\ell i},\;\bar u_\ell,\;p_\ell$ & Unit gate energy, its mean across units and the gate participation ratio.\\
$\hat\kappa_\ell,\;\kappa_\ell(t)$ & Initialization estimate of drive and measured accumulated drive along the initial error direction.\\
$\Lambda_\ell,\;R_\ell$ & Activity--label covariance and its regularized, scale-normalized decoding statistic.\\
$G_\ell^{\rm DFA},\;G_\ell^{\rm BP}$ & DFA gradient estimate ($\Delta W_\ell=-\eta G_\ell^{\rm DFA}$) and true weight gradient on the same probe.\\
$\Pi_\ell,\;\cos\alpha_\ell$ & Normalized signed gradient projection and gradient-direction cosine.\\
$g,\;\eta_{\rm hid},\;\eta_{\rm out}$ & Feedback standard deviation and hidden/output learning rates.\\
$a,\;T_{\rm plateau}$ & Effective hidden rate $a=\eta_{\rm hid}g$ and time from detected plateau onset to exit; scalar $a$ is distinct from $a_\ell$.\\
\bottomrule
\end{longtable}\endgroup

\begin{table}[htbp]\centering\small
\caption{The diagnostics answer different questions. A single value does not establish the properties in the final column.}\label{tab:diagnostic_taxonomy}
\begin{tabular}{>{\raggedright\arraybackslash}p{.23\linewidth}>{\raggedright\arraybackslash}p{.32\linewidth}>{\raggedright\arraybackslash}p{.34\linewidth}}\toprule
Diagnostic & Measures & Does not by itself establish\\\midrule
Hidden cosine & Similarity of activity directions & Equal magnitudes or loss of decodability\\
Gate participation $p_\ell$ & Distribution of squared derivatives across units & Total gate strength or descent\\
Mean gate energy $\bar u_\ell$ & Average squared derivative & Feedback magnitude or alignment\\
Error norm $\|\bar e\|$ & Shared output teaching signal & Delivered parameter displacement\\
Gradient cosine $\cos\alpha_\ell$ & Direction of the gradient estimate relative to BP & Size of loss improvement\\
Projected descent $\Pi_\ell$ & Alignment weighted by relative magnitude & Optimizer-preconditioned descent\\
Weight alignment & Feedback versus downstream weight product & Gradient alignment through gates\\
Label covariance $\|\Lambda_\ell\|_F^2$ & Amplitude of activity--label association & Scale-normalized decodability\\\bottomrule
\end{tabular}\end{table}

\paragraph{Hidden cosine.}
Mean pairwise cosine averages $h(x)^\top h(x')/(\|h(x)\|\|h(x')\|)$ over distinct pairs among 512 probe examples. A value near one indicates similar directions; by itself it does not establish equal vector magnitudes or an input-independent representation. % provenance-ok: diagnostic subsample size

\paragraph{Teaching-signal strength and useful descent.}
Let $G_\ell^{\rm DFA}=\langle\delta_\ell h_{\ell-1}^{\top}\rangle$ and $G_\ell^{\rm BP}=\nabla_{W_\ell}\mathcal L$ on the same probe sub-batch, where $\mathcal L$ denotes its average loss. The signed projected descent, the projected BP-step ratio of \citet{safaai2026}, is
\begin{equation}
 \Pi_\ell=\frac{\langle G_\ell^{\rm DFA},G_\ell^{\rm BP}\rangle_F}{\|G_\ell^{\rm BP}\|_F^2}
 =\frac{\|G_\ell^{\rm DFA}\|_F}{\|G_\ell^{\rm BP}\|_F}\cos\alpha_\ell,
 \label{eq:projection}
\end{equation}
where $\langle U,V\rangle_F=\operatorname{tr}(U^\top V)$ is the Frobenius inner product, $\operatorname{tr}$ sums diagonal entries, and $\cos\alpha_\ell$ is the cosine between the two vectorized matrices. A positive value predicts local loss decrease from that hidden weight update; a negative value predicts increase, to first order in the step size. The statistic includes relative update magnitude and need not lie in $[-1,1]$. These diagnostics use the raw DFA rule; intervention-specific gradients are not reconstructed by this diagnostic. Weight alignment in the teacher--student comparison is the cosine between vectorized $B_\ell$ and $M_\ell=(W_{\rm out}W_L\cdots W_{\ell+1})^\top$, without derivative gates (Appendix~\ref{app:demarcation}).

\paragraph{Label covariance and linear decoding.}
For classification diagnostics, $y$ denotes the original one-hot class target, including in the multi-label and scaled-target controls. Let $\Sigma_{h,\ell}=\Cov(h_\ell,h_\ell)$ and $\Lambda_\ell=\Cov(h_\ell,y)$. The energy $\|\Lambda_\ell\|_F^2$ measures the amplitude of activity--label association. Its reported fold decrease is the ratio of the initial value to the minimum after step 30, computed within each run and then averaged. To remove much of the dependence on activity scale, we use the regularized statistic
\begin{equation}
 R_\ell=\frac{\operatorname{tr}[\Lambda_\ell^\top(\Sigma_{h,\ell}+\epsilon_\ell I)^{-1}\Lambda_\ell]}{C^{-1}\sum_c\operatorname{Var}(y_c)},
 \qquad \epsilon_\ell=10^{-4}\operatorname{tr}(\Sigma_{h,\ell})/d_\ell.
\end{equation}
Here $I$ is the identity, $\epsilon_\ell$ is a ridge regularizer, and $C$ counts the original classes, including when the readout has extra binary outputs. The statistic uses at most 1,024 probe examples, with target-variance normalization from the full probe. It quantifies linearly accessible covariance after correcting for feature scale; it is not held-out decoder accuracy or an information-theoretic quantity. % provenance-ok: diagnostic ridge and subsample size
Because the denominator is the average rather than total target variance, the ideal population statistic can approach $C$, not one. The matched-checkpoint comparison in Table~\ref{tab:E7} evaluates both statistics when covariance energy is smallest among checkpoints with a recorded $R_\ell$; independently sampled minima are not compared.

\paragraph{Centered dimensionality.}
We measure the entropy effective rank of the centered covariance $\Sigma_{h,\ell}$ \citep{roy2007}. If its eigenvalues are $\lambda_j$ and $w_j=\lambda_j/\sum_k\lambda_k$, the effective rank is $\exp(-\sum_j w_j\log w_j)$, with $0\log0=0$. It uses 1,024 probe examples and describes how variance is distributed across activity directions, independently of its total amplitude. Gate participation instead describes how squared activation derivatives are distributed across units. % provenance-ok: diagnostic subsample size

\paragraph{Gate sets.}
For ReLU, an inactive unit has positive preactivation on less than $5\%$ of probe inputs; an always-active unit has positive preactivation on more than $95\%$. The latter need not have constant output. % provenance-ok: ReLU diagnostic thresholds
For each snapshot at step $t$, the gate-energy set $S_t$ contains the top fifth of units ranked by $u_{\ell i}$. Turnover is measured by Jaccard overlap $|S_t\cap S_{120}|/|S_t\cup S_{120}|$. For independent sets containing a fraction $f$ of units, the large-width chance reference is $f/(2-f)$. The Gaussian reconstruction uses each saved $(\mu_{\ell i},\sigma_{\ell i})$ to compute $u_{\ell i}=\E_{Z\sim\mathcal N(\mu_{\ell i},\sigma_{\ell i}^2)}\sech^4Z$, then substitutes these energies into Equation~\eqref{eq:participation}. % provenance-ok: gate-set protocol

\paragraph{Virtual-update diagnostics.}
At fixed training snapshots, we copy the network and apply a full-probe SGD step. Before/after means and every term in the drift budget use the same validation probe; the original network and training optimizer are unchanged. This measures a full-probe local update, not the actual stochastic training step.

For gate-set comparisons, we instead change one hidden layer at a time. We mask rows of its local weight gradient and the corresponding bias entries using all units, the current top fifth by gate energy, the top fifth fixed at step 120, a fixed random fifth, or the fixed set's complement. Each virtual displacement is normalized to joint weight-plus-bias norm $10^{-3}$. We compare its direction with the true probe gradient and record its finite loss change. Current, fixed and random fifths match unit count; the complement is larger. All comparisons match displacement norm. % provenance-ok: mask protocol and virtual-step size

\subsection{Plateau detection and reporting}
The five-point-smoothed probe loss must remain within $3\%$ of the constant-predictor loss for at least 50 steps, with probe accuracy no greater than $\max_c\pi_c+0.05$. Exit is the first later logged step below $0.9\mathcal L_{\rm prior}$. Plateau duration is exit minus onset. Separately, \emph{learning time} is the first crossing below $0.9\mathcal L_{\rm prior}$ that remains below it for 50 updates, whether or not a plateau was detected; this permits the same comparison for all optimizers and controls. Participation minima use the first 300 steps because late saturation can also lower participation. Peak cosine uses the full recorded trajectory unless a figure specifies a shorter window. Curves show means across trajectories; shaded bands, where present, show one standard deviation, not confidence intervals. Trajectory bands use the population standard deviation across runs; tables and error bars use the sample standard deviation. Model timing uses the analogous loss criterion without an accuracy condition. Unless specified otherwise, $r$ denotes Pearson correlation; the unit-drift comparison uses Spearman rank correlation. Model correlations pool condition--seed--layer observations, which are not independent replicates. % provenance-ok: detection protocol

\subsection{Intervention implementations and cost}
\label{app:interventions}
Error centering subtracts the minibatch mean before feedback projection. Teaching-signal centering subtracts the mean after applying derivative gates. In both cases, the output layer still receives its ordinary gradient. Prior-bias initialization sets sigmoid biases to nominal prior logits \citep{lin2017}; the balanced MNIST controls use $\operatorname{logit}(1/C)$. The distinct mean-logit calibration control uses empirical class frequencies and a training subsample to shift initial mean logits toward the corresponding prior logits. Because the sigmoid is nonlinear, this latter operation only approximately calibrates the mean prediction.

Batch normalization is applied to each hidden linear preactivation before the activation function, with trainable scale and offset. The local DFA gradient differentiates through the activation and minibatch normalization; inputs from earlier blocks are detached. Probe evaluation and snapshots also use their own batch statistics, rather than stored running statistics. Reported preactivation moments and derivative gates are measured after normalization and its affine transform; the gate statistic does not include the normalization Jacobian. The trainable offset means that normalized preactivations need not retain zero mean.

The local-error conditioner forms an exponential moving average $S_\ell$ of $\langle\delta_\ell\delta_\ell^\top\rangle$, with decay $0.99$. It replaces each teaching signal by $(S_\ell+\lambda_\ell I)^{-1}\delta_\ell$ and rescales the transformed batch to the original Frobenius norm. Here $I$ is the identity and $\lambda_\ell=\lambda_{\rm rel}\operatorname{tr}(S_\ell)/d_\ell$, where $\lambda_{\rm rel}$ is the dimensionless ridge strength varied in Figure~\ref{fig:cures}d. This is inverse-second-moment preconditioning, not inverse-square-root whitening. It acts in the $d_\ell$-dimensional hidden teaching space, as in the error-side operator of \citet{safaai2026}. % provenance-ok: conditioner decay
Our control adapts that operator: the cited implementation uses current-minibatch moments and absolute additive ridges, whereas ours uses an exponential average, trace-relative damping and restoration of the teaching-signal norm. The comparison should not be read as an identical implementation of the published method.

The orthogonalization control applies five Newton--Schulz iterations to each hidden weight-gradient matrix and restores its original Frobenius norm. It is Muon-style gradient orthogonalization \citep{muon2024}, not the full momentum-based optimizer. Table~\ref{tab:cures} gives both asymptotic overhead and measured runtimes for the implemented controls. % provenance-ok: orthogonalization iteration count

\begin{table}[htbp]
\centering\small
\caption{Additional per-step cost for minibatch size $n$, $C$ outputs and square hidden maps of width $d$, and descriptive mean wall clock per run relative to plain DFA on the base network. Rectangular orthogonalization costs depend on both matrix dimensions. Costs exclude the ordinary DFA update.}
\label{tab:cures}
\begin{tabular}{lllc}\toprule
Intervention & Extra operations & Persistent state & Wall clock \\\midrule
Center output error & $O(nC)$ & none & \numCureCenterEOverhead \\
Center teaching signal & $O(nd)$ per layer & none & \numCureCenterOverhead \\
Prior bias / faster readout & none during training & none & \numCurePriorOverhead\,/\,\numCureOutlrOverhead \\
Batch normalization & $O(nd)$ per layer & $O(d)$ per layer & \numCureBNOverhead \\
Orthogonalize updates & $O(d^3)$ per layer & none beyond SGD & \numCureMuonOverhead \\
Local-error conditioner & $O(nd^2+d^3)$ per layer & $O(d^2)$ per layer & \numCureWhitenOverhead \\\bottomrule
\end{tabular}
\end{table}

The timing ratios use three paired runs and include probes and snapshots as well as training. Their configurations record CUDA, but not the specific GPU model. Small differences near unity should be interpreted as comparable measured cost, rather than established speed improvements. % provenance-ok: timing cohort

\subsection{Code and computation}
\label{app:code}
The supplementary archive contains the training configurations, analysis and figure scripts, tests, and reference outputs for the additional controls. Complete condition summaries and a mapping from each printed comparison to its source are included in \path{results/paper_summary_sources/}; they retain sweeps omitted from the printed supplement. The README gives regeneration commands and identifies which full training logs must be regenerated. Numerical macros and tables are generated from these records.

The frozen-feature data include the fitting/evaluation protocol, and the fresh-architecture data include predictions saved before training. Virtual-update measurements record configurations and probe/feedback hashes. The execution manifest records an A100 MIG slice for the additional GPU controls and CPU integrations for the reduced model; per-run configurations also record device and wall time. The additional learning-rate and imbalanced-calibration runs (E21 and E22) used CPUs. The intervention costs in Table~\ref{tab:cures} are implementation- and hardware-dependent.

\section{Derivations and reduced-model assumptions}
\label{app:derivations}

We derive the exact update decomposition, then state the approximations that yield mean drift, accumulated drive and gate participation. The final subsection extends the model to approximate plateau exit. Empirical checks of the omitted terms and predictions are collected in Appendix~\ref{app:results}.

\subsection{Exact mean--covariance decomposition}
\label{app:decomposition}
For any batch, $\langle\delta h^\top\rangle=\langle(\delta-\bar\delta)(h-\bar h)^\top\rangle+\bar\delta\bar h^\top$. Substituting $\delta=\gamma\odot B(\ebar+\etil)$ and $\langle\etil\rangle=0$ gives Equation~\eqref{eq:decomp}; the bias update follows from Equation~\eqref{eq:dfa}. At initialization, zero-mean Gaussian feedback independent of the forward computation gives the residual zero expectation over feedback draws. This does not bound its norm in one network or during training.

The decomposition by output-error mean is different from the decomposition by teaching-signal mean. Suppressing the layer index, put $v=B\bar e$, and let $\operatorname{diag}(v)$ denote the diagonal matrix with entries $v_i$. At a fixed network state, the weight update contributed by shared output error is exactly
\begin{equation}
 \Delta W_{\bar e}=-\eta\operatorname{diag}(v)\langle\gamma h^\top\rangle
 =-\eta(\bar\gamma\odot v)\bar h^\top
  -\eta\operatorname{diag}(v)\Cov(\gamma,h).
 \label{eq:shared-error-update}
\end{equation}
The first term has rank at most one; the second need not. Pre-gate error centering removes both, while centering the post-gate teaching signal removes its mean outer product and bias update. These operations therefore test different interventions. Appendix~\ref{app:review_closure} measures the two shared-error components alongside the readout budget.

\subsection{Gated drift}
\label{app:drift}
To turn the mean update into a prediction of saturation, consider one layer with trainable biases and suppress its index. Write $\mu_i=w_i^\top\hb+b_i$ and $v=B\ebar$, where $w_i^\top$ is row $i$ of $W$. Holding presynaptic activity fixed, the exact local update is
\begin{equation}
 \Delta\mu_i=-\eta(\bar\gamma_i v_i+r_i)(\|\hb\|^2+1)
 -\eta\Cov(\delta_i,h)\hb.
\end{equation}
The full continuous-time trajectory additionally contains $w_i^\top\dot{\hb}$, where a dot denotes a derivative with respect to update count. Dropping this upstream contribution, the covariance contribution and $r_i$, and setting $\bar\gamma_i\approx\sech^2\mu_i$, gives Equation~\eqref{eq:drift}. Evolving the energies in the reduced model does not restore the omitted terms.

For a finite simultaneous network update, the complete identity includes $W\Delta\hb+\Delta W\Delta\hb$ in addition to the local terms above. We measure all five vector contributions with virtual full-probe SGD updates in Appendix~\ref{app:drift_budget}. This tests the local closure separately from the empirical accuracy of the reduced model.

For a fixed error direction, define $s(t)=\eta\int_0^t\langle\ebar(t'),\hat e_0\rangle(H_{\ell-1}(t')+1)dt'$ and $\hat v=B_\ell\hat e_0$. Separating variables gives
\begin{equation}
 F(\mu_i(t))=F(\mu_i(0))-\hat v_i s(t),
 \qquad F(\mu)=\frac{\mu}{2}+\frac{\sinh(2\mu)}{4}.
 \label{eq:quadrature-map}
\end{equation}
For large $|\hat v_i|s$, $|\mu_i|\approx\tfrac12\ln(8|\hat v_i|s)$. Since $F' (\mu)=\cosh^2\mu$, it integrates the inverse derivative gate. The accumulated drive can therefore grow much faster than the saturated mean. If hidden biases are frozen, the physical drift factor is $H_{\ell-1}$ rather than $H_{\ell-1}+1$.

\subsection{Readout dynamics and the dose}
\label{app:dose}
With $\sigma(z)=(1+e^{-z})^{-1}$, each mean logit in the symmetric mean-only sigmoid model equals $z$, $\ebar=(\sigma(z)-1/C)\mathbf1$, and $\dot z=-\eta_{\rm out}(H_L+1)(\sigma(z)-1/C)$. Starting at $z=0$ with $C>2$, the fixed point is $z^*=\operatorname{logit}(1/C)$, where $\operatorname{logit}(q)=\ln[q/(1-q)]$. For constant $H_L$,
\begin{equation}
 \int_0^\infty\|\ebar\|dt
 =\frac{\sqrt C}{\eta_{\rm out}(H_L+1)}\int_{z^*}^{0}dz
 =\frac{\sqrt C\ln(C-1)}{\eta_{\rm out}(H_L+1)}.
\label{eq:dose}\end{equation}
For $C=2$ there is no initial drive; the effective-slope formula has limiting value $1/4$, while the dose is zero. Near the fixed point, the tail decay rate is $\eta_{\rm out}(H_L+1)\pi(1-\pi)$ for $\pi=1/C$. This tail rate differs from the effective slope that summarizes the whole integral.

At an exactly constant initial prediction $q$, the sigmoid mean-error norm is $\sqrt C|q-1/C|$ for balanced one-hot targets. Uniform softmax instead gives $\|C^{-1}\mathbf1-\pi\|$. Random-readout predictions fluctuate across inputs, so these ideal formulas need not equal the measured initial norm. For general readouts the reduced model integrates the readout equation numerically and uses the actual initial mean logits and target mean.

For arbitrary constant $q\in(0,1)$, the same fixed-activity sigmoid calculation gives total dose proportional to $\sqrt C|\operatorname{logit}(q)-\operatorname{logit}(1/C)|$, rather than to $\sqrt C|q-1/C|$ alone. The product scaling observed when varying $q$ is therefore empirical, not an exact consequence of keeping $\sigma'_{\rm eff}$ fixed.

The online diagnostic uses Equation~\eqref{eq:delivered}, projecting onto the mean-error direction of the first training minibatch. Its code normalizes with a small numerical stabilizer; an exactly zero initial vector therefore gives zero projected dose. This does not rule out later stochastic or residual-driven activity. The logged diagnostic retains the $+1$ convention even in frozen-bias controls; the reduced model removes the bias channel for those controls. For transient comparisons, the sum is stopped at the first logged point where the validation-probe mean-error norm is below $15\%$ of its initial probe value. % provenance-ok: diagnostic cutoff

Finite-batch error fluctuations remain even at zero population mean; Appendix~\ref{app:batch_test} derives and tests their sampling law separately from collapse.

\subsection{Participation quadrature}
\label{app:quadrature}
We now relate the mean shifts to concentration of gate energy across units. With negligible within-unit spread, $u_i=\sech^4\mu_i$. For initially zero means and independent coefficients $\hat v_i\sim\mathcal N(0,g^2)$, Equation~\eqref{eq:quadrature-map} reduces the large-width gate moments to one-dimensional integrals in $\kap=gs$. Here $\bar u$ and $\overline{u^2}$ average over units, and $\varphi$ is the standard normal density. As $\kap\to\infty$, substitution $w=|\hat v_i|s=F(|\mu|)$ gives
\begin{align}
 \bar u&\sim\frac{2\varphi(0)}{\kap}\int_0^\infty\sech^2\mu\,d\mu
 =\frac{\sqrt{2/\pi}}{\kap},\\
 \overline{u^2}&\sim\frac{2\varphi(0)}{\kap}\int_0^\infty\sech^6\mu\,d\mu
 =\frac{8}{15}\frac{\sqrt{2/\pi}}{\kap}.
\end{align}
Their ratio gives $p=\bar u^2/\overline{u^2}\sim(15/8)\sqrt{2/\pi}/\kap$. This is an asymptote of the stated closure, not an exact finite-width identity for trained networks.
The large-width limit precedes the large-dose limit. Since finite-width participation is at least $1/d_\ell$, the asymptote cannot be extrapolated to arbitrarily large dose at fixed width.

\subsection{Complete common-mode reduced model}
\label{app:model}
The implemented model uses a common preactivation variance $\sigma_\ell^2$ within each layer and a separate mean $\mu_{\ell i}$ for each unit. The model replaces the narrow-spread gate in Equation~\eqref{eq:drift} by a Gaussian average. For $Z\sim\mathcal N(0,1)$, it computes mean activity $m_{\ell i}$, mean gate $\bar\gamma_{\ell i}$ and gate energy $u_{\ell i}$:
\begin{equation}
 m_{\ell i}=\E_Z\tanh(\mu_{\ell i}+\sigma_\ell Z),\quad
 \bar\gamma_{\ell i}=\E_Z\sech^2(\mu_{\ell i}+\sigma_\ell Z),\quad
 u_{\ell i}=\E_Z\sech^4(\mu_{\ell i}+\sigma_\ell Z).
\end{equation}
We set $H_\ell=\sum_i m_{\ell i}^2$ and use these gate energies in Equation~\eqref{eq:participation}. The first-layer variance is initialized from the network and held fixed. Subsequent variances follow $\sigma_\ell^2=q_\ell\bar u_{\ell-1}\sigma_{\ell-1}^2$, with $q_\ell=d_{\ell-1}\operatorname{Var}(W_{\ell,ij})$ fixed at initialization. This is a random-weight propagation approximation; it drops feature correlations and later changes in forward-weight statistics.

Let $s_{\rm out}$ be the sigmoid, softmax or identity output map, and $\bar y$ the training target mean. With indicator $\iota_b=1$ for trainable hidden biases and $\iota_b=0$ for frozen biases, the integrated mean equations are
\begin{equation}
 \dot\mu_{\ell i}=-\eta_{\rm hid}\bar\gamma_{\ell i}(B_\ell\ebar)_i(H_{\ell-1}+\iota_b),
 \qquad \dot{\bar z}=-\eta_{\rm out}(H_L+1)\ebar,
 \qquad \ebar=s_{\rm out}(\bar z)-\bar y.
\end{equation}
Applying the output map to the mean logit is another closure: generally $\E s_{\rm out}(z)\ne s_{\rm out}(\E z)$. All initial means, first-layer spread, feedback matrices and weight variances are read from the initialized network. No trajectory parameters are fitted. The implementation uses 64-node Gauss--Hermite quadrature and forward Euler integration, with one substep per training update by default. Appendix~\ref{app:resolution} checks sensitivity to both numerical resolutions. % provenance-ok: numerical integration protocol

The model's predicted hidden cosine is $H_\ell/(H_\ell+d_\ell\bar u_\ell\sigma_\ell^2)$. It approximates the mean pairwise cosine by a ratio of expected inner products and squared norms, requiring concentration of activity norms. It is not an exact cosine identity. In particular, high cosine alone cannot distinguish small centered variation from a large common offset.

Appendix~\ref{app:model_validation} evaluates these predictions across settings and reports detection failures.

\subsection{Approximate covariance dynamics and exit prediction}
\label{app:escape}
To estimate when loss starts falling below the prior, we augment the collapse model with scalar covariance amplitudes $A_\ell\approx\|\Lambda_\ell\|_F$. Define $A_0=\|\Cov(x,y)\|_F$, $\rho_\ell=d_\ell\operatorname{Var}(W_{\ell,ij})$, and an accumulated local contribution $J_\ell$ initialized to zero. The implementation uses
\begin{equation}
 A_\ell=\sqrt{\rho_\ell\bar u_\ell}\,A_{\ell-1}+J_\ell,
 \qquad \dot J_\ell=\eta_{\rm hid}g\sqrt{d_\ell/C}\,
 \frac{\bar u_\ell}{\sqrt{p_\ell}}A_{\ell-1}^2.
 \label{eq:escape-model}
\end{equation}
In Equation~\eqref{eq:escape-model}, the first term transmits upstream covariance instantaneously; the second accumulates a local learning contribution. Adding their norms is a phenomenological closure, not an exact matrix-covariance identity. The accumulated contribution is not attenuated again when gates later change.

An accumulated loss-reduction variable $E$ starts at zero and evolves as $\dot E=\eta_{\rm out}A_L^2$. The model loss is the mean-only readout loss minus $E$, with a constant offset chosen so that the mean-only fixed-point loss equals the data's prior-predictor loss. Near a constant predictor, $\Cov(e,h_L)\approx-\Lambda_L^\top$, motivating this descent estimate. Away from that regime, prediction-dependent covariance and higher-order terms are omitted. The integration is used to estimate the first exit below $0.9\mathcal L_{\rm prior}$, not to model the eventual training loss. Because the variance cascade lacks recovery dynamics, the model can predict an exit while its gates remain concentrated. It therefore cannot account for the observed participation rebound. % provenance-ok: exit criterion

\section{Tests of the mechanism and reduced model}
\label{app:results}\label{app:tables}
We first test the factors that produce the shared drive, then measure the terms omitted by the model. Prediction accuracy is assessed separately from these local checks, with all model detection failures retained.

\subsection{Mean-drive and alignment controls}
\label{app:alignment_controls}
\begin{figure*}[t]\centering\includegraphics[width=\textwidth]{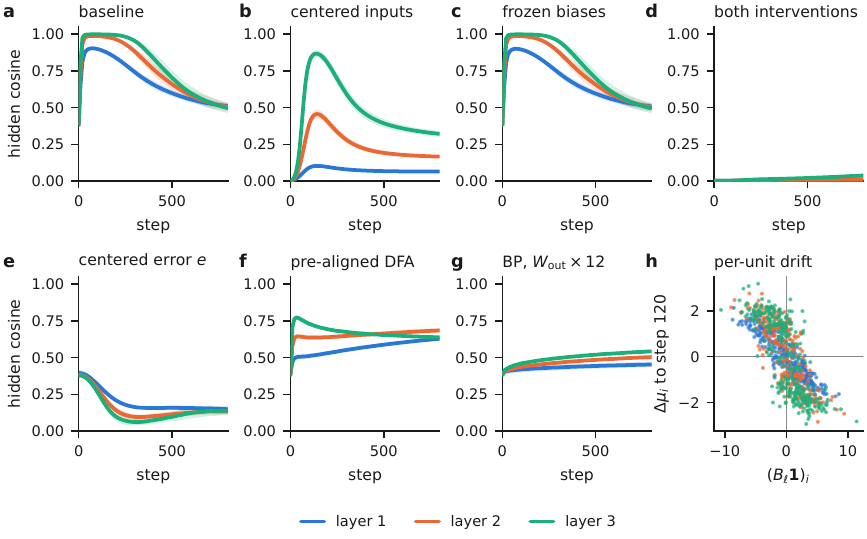}
\caption{\textbf{Causal dissection.} (a--d) Input centering $\times$ frozen hidden biases: the combined intervention prevents collapse at every layer without forcing all hidden means to zero. (e) Output error centered before projection. (f) DFA with feedback pre-aligned to the transposed downstream map at the Frobenius norm of the random draw. (g) BP with its forward readout weights scaled by 12, which also raises $\|\ebar_0\|$ to \numBPxEbar: neither alignment control collapses. (h) Unit mean changes over the first 120 steps versus $(B_\ell\mathbf1)_i$, for one baseline trajectory (seeds 0,0). Colours denote layers. Curves in a--g average runs; bands show one standard deviation. Counts are in Table~\ref{tab:E1}.}\label{fig:dissection}\end{figure*}

The centering controls test the two routes through which a mean teaching signal shifts a unit's preactivation. Removing the input mean protects the first layer, but deeper collapse persists; freezing hidden biases alone is also insufficient. Together they prevent collapse throughout the network (Figure~\ref{fig:dissection}a--d). Global standardization instead raises the MNIST input mean energy and deepens layer-1 gate concentration (minimum $p_1$ of \numStdLOneP\ versus \numRawLOneP), although peak cosine is lower (\numStdLOneCos\ versus \numRawLOneCos). Centering output error acts on the other factor and suppresses baseline collapse (panel~e). Table~\ref{tab:E1} summarizes these controls; values are means and standard deviations across trajectories, and unavailable endpoints remain marked.
% Generated by scripts/supplement_tables.py.
\begin{table}[t]\centering\footnotesize\setlength{\tabcolsep}{3.5pt}
\caption{MNIST baseline and mean-drive controls. Peak cosine uses the recorded trajectory; participation uses the first 300 updates. Accuracy is measured at step 3000. The frozen-readout control ends at step 1500, so its endpoint is unavailable (--). Softmax rows use globally standardized inputs, while other rows use raw pixels unless input centering is specified. Values are means $\pm$ standard deviations.}\label{tab:E1}
\begin{tabular}{lrrrr}\toprule
Condition & $n$ & Peak $\cos_3$ & Min. $p_3$ & Accuracy\\\midrule
Plain DFA & 15 & $0.998\pm0.000$ & $0.225\pm0.029$ & $0.894\pm0.006$\\
Backpropagation & 5 & $0.731\pm0.012$ & $0.951\pm0.004$ & $0.861\pm0.006$\\
Layerwise FA & 15 & $0.991\pm0.002$ & $0.342\pm0.039$ & $0.882\pm0.008$\\
Pre-aligned DFA & 15 & $0.772\pm0.015$ & $0.882\pm0.007$ & $0.912\pm0.004$\\
BP, output weights $\times12$ & 5 & $0.607\pm0.015$ & $0.986\pm0.002$ & $0.927\pm0.004$\\
Center output error & 15 & $0.379\pm0.017$ & $0.932\pm0.011$ & $0.877\pm0.007$\\
Center teaching signal & 15 & $0.521\pm0.062$ & $0.743\pm0.132$ & $0.861\pm0.007$\\
Center inputs & 15 & $0.867\pm0.024$ & $0.704\pm0.030$ & $0.897\pm0.006$\\
Freeze hidden biases & 15 & $0.998\pm0.001$ & $0.229\pm0.029$ & $0.894\pm0.006$\\
Center inputs + freeze biases & 15 & $0.081\pm0.011$ & $0.922\pm0.015$ & $0.893\pm0.007$\\
Faster readout & 15 & $0.923\pm0.010$ & $0.816\pm0.020$ & $0.911\pm0.006$\\
Frozen readout & 15 & $1.000\pm0.000$ & $0.018\pm0.006$ & --\\
Prior-bias initialization & 15 & $0.474\pm0.145$ & $0.928\pm0.065$ & $0.897\pm0.006$\\
Balanced softmax & 15 & $0.271\pm0.032$ & $0.929\pm0.009$ & $0.922\pm0.004$\\
Imbalanced softmax & 15 & $0.938\pm0.028$ & $0.477\pm0.085$ & $0.960\pm0.005$\\
Imbalanced softmax, BP & 5 & $0.386\pm0.021$ & $0.972\pm0.002$ & $0.947\pm0.005$\\
\bottomrule\end{tabular}\end{table}

% Generated by scripts/submission_reporting.py.
\begin{table}[t]\centering\scriptsize\setlength{\tabcolsep}{3.5pt}
\caption{Learning cost of collapse in the MNIST baseline cohort. Learning time is the first probe loss below 90\% of the constant-predictor loss that stays below it for 50 updates (the criterion of Table~\ref{tab:review_timing}); mean $\pm$ standard deviation. Accuracy columns give the median first step reaching 50\% and 80\% probe accuracy; the last column is accuracy at step 3000. Peak cosine uses the full trajectory, as in Table~\ref{tab:E1}. Centering prevents collapse and speeds discrimination but lowers mean-activity energy, which slows the readout's fit to the class prior; the sigmoid loss then stays high although accuracy rises.}\label{tab:submission_cost}
\begin{tabular}{lrrrrrr}
\toprule
Condition & $n$ & Peak $\cos_3$ & Learning time & 50\% acc. & 80\% acc. & Accuracy\\\midrule
Plain DFA & 15 & $0.998\pm0.000$ & $531\pm34$ & 690 & 1220 & $0.894\pm0.006$\\
Frozen hidden biases & 15 & $0.998\pm0.001$ & $528\pm35$ & 680 & 1220 & $0.894\pm0.006$\\
Centered inputs & 15 & $0.867\pm0.024$ & $345\pm35$ & 430 & 910 & $0.897\pm0.006$\\
Centered inputs + frozen biases & 15 & $0.081\pm0.011$ & $1153\pm30$ & 370 & 1050 & $0.893\pm0.007$\\
Centered output error & 15 & $0.379\pm0.017$ & $1898\pm153$ & 370 & 1250 & $0.877\pm0.007$\\
Prior-bias initialization & 15 & $0.474\pm0.145$ & $288\pm35$ & 420 & 860 & $0.897\pm0.006$\\
Faster readout ($10\times$) & 15 & $0.923\pm0.010$ & $152\pm11$ & 180 & 460 & $0.911\pm0.006$\\
Pre-aligned DFA & 15 & $0.772\pm0.015$ & $22\pm4$ & 20 & 100 & $0.912\pm0.004$\\
Backpropagation & 5 & $0.731\pm0.012$ & $234\pm17$ & 210 & 1060 & $0.861\pm0.006$\\
\bottomrule\end{tabular}\end{table}

Feedback orientation provides a separate test. Pre-aligned DFA learns quickly with little collapse (panel~f), although matching the feedback matrix's Frobenius norm does not preserve $B_\ell\bar e$. BP remains uncollapsed when its output weights are amplified (panel~g), so its smaller initial feedback scale is insufficient to explain the difference. Ten updates into training, the ratio of shared to input-dependent error norms is \numSelfTermBase\ under random feedback, \numSelfTermAligned\ with pre-aligned feedback, \numSelfTermBPx\ for BP with a twelvefold readout, and \numSelfTermBP\ for unscaled BP. These measurements are consistent with the local restoring contribution in Equation~\eqref{eq:selfterm}, although neither alignment control isolates it. Unit mean shifts are anticorrelated with $(B_\ell\mathbf1)_i$, as the negative sign in Equation~\eqref{eq:drift} predicts (panel~h). Table~\ref{tab:review_unit_drift} checks this relation across all baseline trajectories. These comparisons support mean-driven saturation while showing that its consequences depend on the direction and speed of input-dependent learning.
% Generated by scripts/review_reporting.py.
\begin{table}[t]\centering\scriptsize\setlength{\tabcolsep}{3.5pt}
\caption{Unit mean shifts from initialization to step 120 against the shared-error feedback coefficient. Values are means and standard deviations of per-run Spearman correlations across the 15 baseline trajectories (five initializations crossed with three feedback draws); they are descriptive spreads, not independent-run confidence intervals.}\label{tab:review_unit_drift}
\begin{tabular}{lrr}
\toprule
Layer & Trajectories & Spearman correlation\\\midrule
1 & 15 & $-0.92\pm0.03$\\
2 & 15 & $-0.86\pm0.03$\\
3 & 15 & $-0.83\pm0.04$\\
\bottomrule\end{tabular}\end{table}

\paragraph{Centered geometry.}\label{app:centered}
High mean pairwise cosine could reflect a shared offset rather than suppressed variation. The centered activity variance $\E_x\|h_3-\hb_3\|^2$ separates the two. Averaging per-run normalized minima within 500 updates gives \numCenteredVarBase\ in the baseline, \numCenteredVarBP\ for BP and \numCenteredVarPrior\ with prior-bias initialization (Table~\ref{tab:submission_centered}). Both error centering and input centering with frozen biases preserve the initial variance over this window. These are averages of per-run minima, not lower bounds for every trajectory. The effective rank of centered activity behaves differently (Appendix~\ref{app:diagnostics}). It falls from \numCenteredRankInit\ to about five under random-feedback DFA whether or not collapse occurs (\numCenteredRankBase\ in the baseline, \numCenteredRankCenterE\ with error centering and \numCenteredRankPrior\ with prior bias), but only to \numCenteredRankBP\ under BP. Compression thus persists without the strong loss of activity amplitude. A low effective rank alone does not identify the common-mode transient.
% Generated by scripts/submission_reporting.py.
\begin{table}[t]\centering\scriptsize\setlength{\tabcolsep}{3.5pt}
\caption{Centered geometry of top-layer activity in the MNIST baseline cohort. The centered variance is $\E_x\|h_3-\hb_3\|^2$, the amplitude of activity differences across inputs, and its minimum within the first 500 updates is given relative to initialization. The centered effective rank is the exponential of the entropy of the normalized centered covariance spectrum; its minimum uses the same 500-update window. Means $\pm$ standard deviations across runs.}\label{tab:submission_centered}
\begin{tabular}{lrrrr}
\toprule
Condition & $n$ & Min.\ variance / initial & Initial rank & Min.\ rank\\\midrule
Plain DFA & 15 & $0.028\pm0.007$ & $45.6\pm0.7$ & $4.3\pm0.3$\\
Backpropagation & 5 & $0.826\pm0.008$ & $45.6\pm0.7$ & $37.4\pm0.6$\\
Pre-aligned DFA & 15 & $0.808\pm0.013$ & $45.6\pm0.7$ & $12.2\pm0.2$\\
Prior-bias initialization & 15 & $0.967\pm0.126$ & $45.6\pm0.7$ & $5.3\pm0.2$\\
Centered inputs + frozen biases & 15 & $1.000\pm0.000$ & $43.7\pm0.6$ & $4.8\pm0.8$\\
Centered output error & 15 & $1.000\pm0.000$ & $45.6\pm0.7$ & $4.6\pm0.6$\\
\bottomrule\end{tabular}\end{table}

\subsection{Scaling with drive strength and learning rate}
\label{app:scaling}
The learning-rate sweep tests whether the drive's strength and duration have separable effects. When both SGD rates vary together, peak hidden cosine (\numEtaCosRange) and participation minima (\numEtaPRange) vary within seed noise, while the rescaled exit time $\eta t_{\rm exit}$ lies in \numEtaExitProduct. These measurements support a change in timescale with little change in collapse severity.

Larger plain-SGD rates, with equal hidden and readout rates, test whether the stall's cost disappears when learning is fast (Table~\ref{tab:submission_lr}). No run diverged up to $\eta=0.1$. Collapse deepens at the largest rates, and the relative saving from prior-bias initialization grows: the time to 80\% accuracy falls by \numLrSavingLow\% at $\eta=10^{-3}$ and by \numLrSavingHigh\% at $\eta=0.1$, where plain DFA needs \numLrDFAHighEighty\ updates, calibrated DFA \numLrPriorHighEighty\ and BP \numLrBPHighEighty. The absolute delay is smaller than at the baseline rate, but a substantial advantage of prior-bias initialization remains. The initial biases use the nominal balanced prior, $q=1/C$, rather than estimating each empirical class frequency. % provenance-ok: learning rates and accuracy threshold
% Generated by scripts/submission_reporting.py.
\begin{table}[t]\centering\scriptsize\setlength{\tabcolsep}{3.5pt}
\caption{Plain-SGD learning rate in the MNIST baseline, with equal hidden and readout rates. Columns give diverged runs (non-finite loss, or final loss above the initial loss), the participation minimum within 300 updates, the median first update reaching 50\% and 80\% probe accuracy, and accuracy after 3,000 updates, over non-diverged runs with seed pairs $(0,0)$, $(1,1)$ and $(2,2)$. The $10^{-3}$ rows reuse the headline runs with initialization seeds 0--2 (BP feedback seeds are unused). Prior bias sets all sigmoid biases to $\operatorname{logit}(1/C)$. Probes are logged every ten updates at $10^{-3}$ and every two at larger rates. Each new run completes 3,000 updates.}\label{tab:submission_lr}
\begin{tabular}{llrrrrrr}
\toprule
$\eta$ & Rule & $n$ & Diverged & Min.\ $p_3$ & 50\% acc. & 80\% acc. & Accuracy\\\midrule
$0.001$ & DFA & 3 & 0 & $0.2143\pm0.0191$ & 690 & 1150 & $0.902\pm0.005$\\
$0.001$ & DFA, prior bias & 3 & 0 & $0.8545\pm0.1410$ & 420 & 800 & $0.904\pm0.003$\\
$0.001$ & BP & 3 & 0 & $0.9504\pm0.0054$ & 210 & 1040 & $0.863\pm0.006$\\
$0.003$ & DFA & 3 & 0 & $0.2053\pm0.0203$ & 236 & 406 & $0.925\pm0.005$\\
$0.003$ & DFA, prior bias & 3 & 0 & $0.8471\pm0.1342$ & 138 & 270 & $0.928\pm0.004$\\
$0.003$ & BP & 3 & 0 & $0.9481\pm0.0058$ & 66 & 324 & $0.904\pm0.004$\\
$0.01$ & DFA & 3 & 0 & $0.1614\pm0.0242$ & 80 & 132 & $0.954\pm0.001$\\
$0.01$ & DFA, prior bias & 3 & 0 & $0.8173\pm0.1341$ & 42 & 82 & $0.954\pm0.001$\\
$0.01$ & BP & 3 & 0 & $0.9296\pm0.0058$ & 22 & 108 & $0.928\pm0.004$\\
$0.03$ & DFA & 3 & 0 & $0.0623\pm0.0105$ & 40 & 66 & $0.967\pm0.003$\\
$0.03$ & DFA, prior bias & 3 & 0 & $0.7554\pm0.1711$ & 16 & 36 & $0.968\pm0.003$\\
$0.03$ & BP & 3 & 0 & $0.8981\pm0.0082$ & 10 & 40 & $0.958\pm0.002$\\
$0.1$ & DFA & 3 & 0 & $0.0038\pm0.0004$ & 104 & 130 & $0.971\pm0.003$\\
$0.1$ & DFA, prior bias & 3 & 0 & $0.5043\pm0.2070$ & 14 & 28 & $0.974\pm0.003$\\
$0.1$ & BP & 3 & 0 & $0.8485\pm0.0097$ & 6 & 20 & $0.974\pm0.003$\\
\bottomrule\end{tabular}\end{table}

Varying gain over \numGainRange\ and initial sigmoid probability gives a common empirical trend against $g\|\ebar_0\|$. Learning time falls as gain rises even as collapse deepens: from \numGainLowTime\ updates at $g=0.03$, with minimum participation \numGainLowP, to \numGainHighTime\ at $g=10$, with \numGainHighP. % provenance-ok: gain values
Across \numMasterN\ distinct runs spanning \numMasterRange, the fitted relation is
\begin{equation}
 1-\max_t\cos_L=\numProductLawPrefactor(g\|\ebar_0\|)^{\numProductLawSlope},
\end{equation}
with log--log correlation \numProductLawR. Changing the initial prediction also changes the effective output slope, so this fit is not a universal consequence of the initialization estimate. For the \numKappaPN\ uncentered runs with $\kap_L>3$, the median product $\kap_L\min_t p_L$ is \numKappaP, compared with the large-dose quadrature limit of $1.50$. % provenance-ok: asymptote and dose threshold

Table~\ref{tab:E2} separately checks the initial-error formula across class counts. The binary case has no drive at exactly uniform prediction, but the random readout leaves a small nonzero mean error.
% Generated by scripts/supplement_tables.py.
\begin{table}[t]\centering\footnotesize\setlength{\tabcolsep}{3.5pt}
\caption{Initial mean error under sigmoid outputs as the number of MNIST classes changes. The ideal value assumes a constant prediction of one half. The measured value uses a random readout. Class restriction also changes training-set size; this comparison tests initialization, not later learning speed.}\label{tab:E2}
\begin{tabular}{lrrr}\toprule
$C$ & $n$ & Ideal $\|\bar e_0\|$ & Measured $\|\bar e_0\|$\\\midrule
2 & 3 & $0.000$ & $0.057\pm0.013$\\
3 & 3 & $0.289$ & $0.266\pm0.088$\\
5 & 3 & $0.671$ & $0.700\pm0.083$\\
10 & 3 & $1.265$ & $1.289\pm0.055$\\
\bottomrule\end{tabular}\end{table}

\subsection{Measured limits of the mean-drive approximation}
\label{app:residual}\label{app:drift_budget}\label{app:review_closure}
The exact decomposition does not guarantee that its residual and upstream terms are small. We measure them at fixed checkpoints using virtual full-probe SGD updates, as defined in Appendix~\ref{app:diagnostics}. All before/after quantities use the same validation examples, and the original training trajectory is unchanged.

\paragraph{Gate--error residual.}
We evaluate the residual after 0, 50, 100, 200 and 300 updates on a fixed 2{,}048-example validation probe (sampler seed 999), using initialization--feedback pairs $(0,0)$, $(1,1)$ and $(2,2)$. Baseline and ReLU runs use MNIST; the CIFAR-10 control uses a tanh MLP with the base protocol otherwise unchanged. Table~\ref{tab:residual_replication} reports ratios to the leading mean-teaching term across all three layers. The residual is small during initial tanh collapse but becomes comparable later, as the leading error decreases. Its relative contribution is larger for ReLU. Absolute norms are retained with the measurements, since a large ratio can result from a small denominator. % provenance-ok: residual measurement protocol
% Generated from results/revision_dynamics/residual.csv.
\begin{table}[t]\centering\small
\caption{Gate--error residual norm relative to the leading mean-teaching norm, mean $\pm$ standard deviation over three seed pairs. Absolute norms are retained in the supplementary data.}\label{tab:residual_replication}
\begin{tabular}{llrrr}\toprule Setting & Step & Layer 1 & Layer 2 & Layer 3\\\midrule
MNIST tanh & 0 & $0.01\pm0.00$ & $0.01\pm0.00$ & $0.01\pm0.00$ \\
 & 50 & $0.22\pm0.02$ & $0.23\pm0.02$ & $0.11\pm0.03$ \\
 & 100 & $1.00\pm0.10$ & $0.83\pm0.09$ & $0.37\pm0.09$ \\
 & 300 & $1.89\pm0.93$ & $1.57\pm0.82$ & $1.48\pm0.89$ \\
\addlinespace
MNIST ReLU & 0 & $0.07\pm0.00$ & $0.06\pm0.01$ & $0.05\pm0.00$ \\
 & 50 & $0.94\pm0.13$ & $0.42\pm0.05$ & $0.19\pm0.07$ \\
 & 100 & $1.11\pm0.13$ & $0.41\pm0.01$ & $0.14\pm0.05$ \\
 & 300 & $1.13\pm0.43$ & $0.74\pm0.29$ & $0.57\pm0.22$ \\
\addlinespace
CIFAR-10 tanh & 0 & $0.02\pm0.00$ & $0.01\pm0.00$ & $0.01\pm0.00$ \\
 & 50 & $0.43\pm0.12$ & $0.06\pm0.00$ & $0.02\pm0.00$ \\
 & 100 & $1.06\pm0.42$ & $0.24\pm0.04$ & $0.07\pm0.02$ \\
 & 300 & $1.75\pm0.64$ & $1.09\pm0.20$ & $0.62\pm0.26$ \\
\addlinespace
\bottomrule\end{tabular}\end{table}

\paragraph{Hidden mean drift.}
For the SGD rule in Equation~\eqref{eq:dfa}, write $\mu_\ell=\E_x a_\ell$ for the vector of unit means. The virtual full-probe updates in Appendix~\ref{app:diagnostics} give the exact finite-step decomposition:
\begin{align}
\Delta\mu_\ell={}&-\eta(H_{\ell-1}+1)\bar\gamma_\ell\odot B_\ell\bar e
-\eta(H_{\ell-1}+1)r_\ell
-\eta\operatorname{Cov}(\delta_\ell,h_{\ell-1})\bar h_{\ell-1}\nonumber\\
&+W_\ell\Delta\bar h_{\ell-1}+\Delta W_\ell\Delta\bar h_{\ell-1}.
\end{align}
All unincremented quantities are evaluated before the step; $\Delta\bar h_{\ell-1}$ is the change in mean presynaptic activity. The first three terms are local, the fourth propagates upstream changes, and the fifth is their finite-step interaction. Their vector sum reconstructs the measured shift to floating-point precision; their norms need not sum.

\begin{figure*}[t]\centering\includegraphics[width=\textwidth]{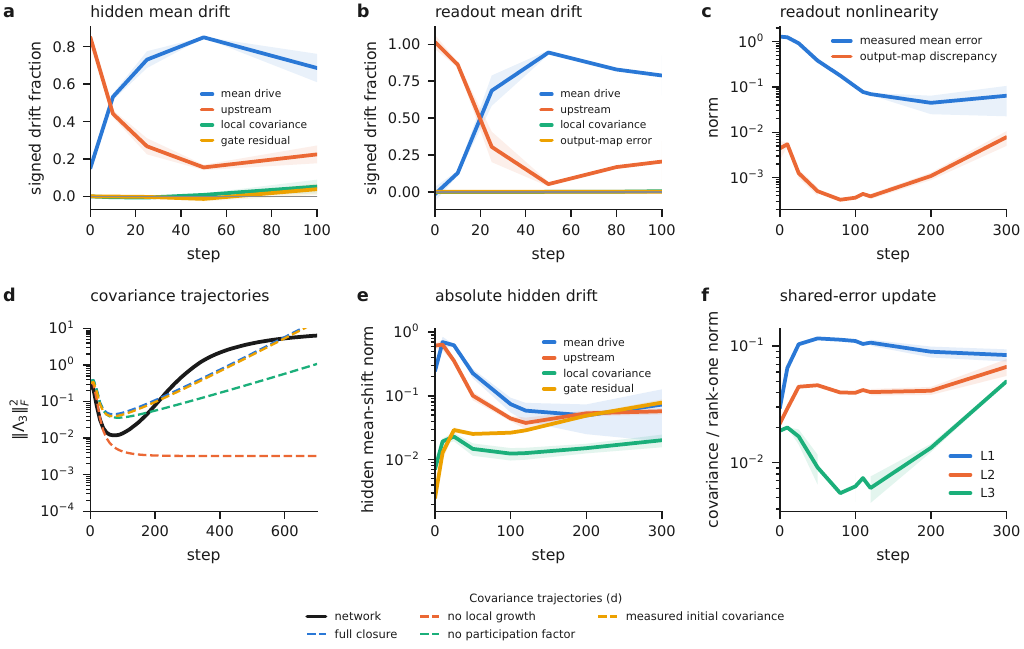}
\caption{\textbf{Checks of the collapse and recovery approximations.} (a,b) Signed contributions along the measured virtual displacement of hidden layer-3 means and readout mean logits; finite cross terms are retained in the data. (c) Measured mean-error norm and the discrepancy from applying the sigmoid to the mean logit. (d) Layer-3 covariance energy in the network and the four recovery approximations defined in Appendix~\ref{app:recovery_tests} (legend below). (e) Absolute norms of the hidden layer-3 drift contributions in a. (f) Norm ratio of the covariance and rank-one contributions of shared output error in Equation~\eqref{eq:shared-error-update}. All panels use three baseline seed pairs; curves show means, and bands where shown one standard deviation. These are snapshot checks and model comparisons, not independent causal interventions.}\label{fig:review_validation}\end{figure*}

A large omitted vector need not explain a large fraction of net motion. For each contribution $d_k$ to $\Delta\mu$, we therefore also report the signed fraction $\langle d_k,\Delta\mu\rangle/\|\Delta\mu\|^2$. These fractions can be negative or exceed one when terms cancel; the complete decomposition sums to one. Upstream motion dominates initially, whereas the leading mean-error term dominates during the collapse (Figure~\ref{fig:review_validation}a). Panel~e gives the corresponding absolute norms; Table~\ref{tab:review_drift} includes the norm of the net displacement. Upstream motion may itself originate in mean-driven updates, so this comparison does not identify independent causal sources.

\paragraph{Readout calibration.}
Readout adaptation helps limit the duration of the drive, so we check it separately. Put $\bar z=W_{\rm out}\bar h_L+b_{\rm out}$ and $e_{\rm cl}=\sigma(\bar z)-\bar y$, the mean-logit approximation to $\bar e$. The same virtual full-probe step gives
\begin{align}
\Delta\bar z={}&-\eta_{\rm out}(H_L+1)e_{\rm cl}
-\eta_{\rm out}(H_L+1)(\bar e-e_{\rm cl})
-\eta_{\rm out}\Cov(e,h_L)\bar h_L\nonumber\\
&+W_{\rm out}\Delta\bar h_L+\Delta W_{\rm out}\Delta\bar h_L.
\end{align}
The second term measures the output-nonlinearity approximation, the third error--activity covariance, and the last two feature motion and its finite-step interaction. Together they reconstruct the measured mean-logit shift. Upstream motion is substantial initially and again later; the leading mean-route fraction is \numReviewReadoutMeanFifty\ at step 50 (Figure~\ref{fig:review_validation}b). The direct output-map discrepancy is much smaller than the early mean error (panel~c).
% Generated by scripts/review_reporting.py.
\begin{table}[t]\centering\scriptsize\setlength{\tabcolsep}{3.5pt}
\caption{Signed projections onto the measured virtual mean displacement, divided by its squared norm. The complete components sum to one; the displayed columns omit the gate residual or output-map error and the finite cross term, so they need not sum to one. Absolute displacement norms are shown because large fractions can arise when net motion is small. Means and standard deviations use three seeds.}\label{tab:review_drift}
\begin{tabular}{lrrrrr}
\toprule
Location & Step & Mean term & Upstream & Covariance & $\|\Delta\mu\|$ or $\|\Delta\bar z\|$\\\midrule
Hidden layer 3 & 0 & $0.15\pm0.01$ & $0.85\pm0.01$ & $-0.00\pm0.00$ & $0.6715\pm0.0644$\\
Hidden layer 3 & 50 & $0.85\pm0.00$ & $0.15\pm0.01$ & $0.01\pm0.01$ & $0.2405\pm0.0426$\\
Hidden layer 3 & 100 & $0.69\pm0.08$ & $0.22\pm0.05$ & $0.05\pm0.04$ & $0.0837\pm0.0187$\\
Hidden layer 3 & 300 & $0.30\pm0.38$ & $0.35\pm0.15$ & $-0.02\pm0.05$ & $0.1083\pm0.0242$\\
Readout & 0 & $-0.02\pm0.05$ & $1.01\pm0.05$ & $-0.00\pm0.00$ & $0.1611\pm0.0378$\\
Readout & 50 & $0.94\pm0.02$ & $0.05\pm0.02$ & $0.00\pm0.00$ & $0.0778\pm0.0082$\\
Readout & 100 & $0.79\pm0.13$ & $0.21\pm0.14$ & $0.00\pm0.00$ & $0.0248\pm0.0031$\\
Readout & 300 & $0.13\pm0.30$ & $0.85\pm0.36$ & $0.02\pm0.01$ & $0.0162\pm0.0052$\\
\bottomrule\end{tabular}\end{table}

Finally, panel~f measures the covariance contribution of shared output error relative to its rank-one contribution in Equation~\eqref{eq:shared-error-update}. The former is not generally zero. Error centering therefore tests removal of shared output error, rather than isolating the rank-one term. Together, these checks support an effective description of early collapse without establishing uniformly negligible omitted terms.

\subsection{Reduced-model accuracy and detection limits}
\label{app:model_validation}
Model agreement must be assessed across distinct physical settings and must include detection failures. We therefore identify each condition by its training parameters, normalize equivalent defaults, and keep initialization and feedback seeds as separate replication indices. Run labels and logging choices do not define new conditions. The comparison uses \numModelConds\ distinct settings and three paired seeds per setting, giving \numCrossSeedRuns\ trajectories without fitted parameters (Figure~\ref{fig:reduced}d,e).

Table~\ref{tab:model_coverage} reports whether the network and model criteria detect a plateau in each trajectory (Appendix~\ref{app:protocol}). This is distinct from detecting representation collapse. Timing agreement is conditional on detecting a plateau in both trajectories; missed and extra plateaus are therefore reported separately. Onsets at zero remain in timing comparisons but cannot enter relative onset errors.
% Generated by scripts/review_reporting.py.
\begin{table}[t]\centering\scriptsize\setlength{\tabcolsep}{3.5pt}
\caption{Participation prediction across distinct physical conditions. Intervals resample whole conditions, retaining their seed and layer observations (2,000 bootstrap draws). Condition means average over seeds and layers; deeper layers contain only two conditions and have no reported interval.}\label{tab:review_model_stats}
\begin{tabular}{lrrrrr}
\toprule
Summary & Conditions & Points & $r$ & 95\% interval & MAE\\\midrule
Pooled & 48 & 450 & 0.89 & $[0.81,0.94]$ & 0.105\\
Condition means & 48 & 48 & 0.91 & $[0.83,0.97]$ & 0.101\\
Layer 1 & 48 & 144 & 0.93 & $[0.87,0.96]$ & 0.128\\
Layer 2 & 48 & 144 & 0.90 & $[0.82,0.95]$ & 0.112\\
Layer 3 & 48 & 144 & 0.86 & $[0.73,0.95]$ & 0.087\\
Layer 4 & 2 & 6 & -0.16 & -- & 0.027\\
Layer 5 & 2 & 6 & 0.41 & -- & 0.010\\
Layer 6 & 2 & 6 & 0.89 & -- & 0.015\\
\bottomrule\end{tabular}\end{table}

% Generated by scripts/submission_reporting.py.
\begin{table}[t]\centering\scriptsize\setlength{\tabcolsep}{3.5pt}
\caption{Participation minima predicted by the reduced model and by the closed-form initialization estimate, $\min(1,1.5/\hat\kappa_\ell)$ with $\hat\kappa_\ell$ from Equation~\eqref{eq:kappa}, on condition--seed--layer points. The closed form uses the sigmoid effective slope; the second block restricts both predictors to the settings its derivation covers. Condition means average seeds and layers, as in Table~\ref{tab:review_model_stats}. Within single-parameter sweeps, the reduced-model correlation over setting-by-layer means is: gain 0.98, output bias 0.99, learning rate 0.99, width 0.98, batch size 0.99, class count 0.10; the class-count sweep includes the two-class setting, where the model predicts almost no drive.}\label{tab:submission_null}
\begin{tabular}{llrrrrr}
\toprule
Settings & Predictor & Points & $r$ & MAE & $r$, means & MAE, means\\\midrule
All 48 settings & Reduced model & 450 & 0.89 & 0.105 & 0.91 & 0.101\\
 & Closed-form estimate & 450 & 0.39 & 0.333 & 0.64 & 0.307\\
37 tanh, sigmoid, $C>2$ settings & Reduced model & 351 & 0.96 & 0.077 & 0.99 & 0.072\\
 & Closed-form estimate & 351 & 0.43 & 0.318 & 0.67 & 0.293\\
\bottomrule\end{tabular}\end{table}

% Generated from reduced_crossseed.json and reduced_grid.json.
\begin{table}[t]\centering\scriptsize
\caption{Plateau detection and timing coverage. Both, network only, model only and neither partition all trajectories by detected onset. Censored model exits are included in Both but excluded from numerical exit errors. Plateau detection uses loss, not hidden-state collapse.}\label{tab:model_coverage}
\begin{tabular}{lrrrrrrr}\toprule Comparison & Runs & Both & Network only & Model only & Neither & Exit pairs & Censored\\\midrule
Parameter sweeps & 144 & 77 & 1 & 13 & 53 & 76 & 1\\
Width--depth grid & 48 & 39 & 4 & 3 & 2 & 39 & 0\\
\bottomrule\end{tabular}\end{table}

For the gain, output-bias, learning-rate, width and batch-size sweeps, correlations between predicted and measured participation minima over setting-by-layer means range from $\numSweepRMin$ to $\numSweepRMax$. The class-count sweep is an exception ($r=\numSweepRClass$); it includes the two-class setting, where the model predicts almost no drive. On the \numValidConds\ settings covered by the closed-form estimate, correlations between condition means, each averaged over seeds and layers, are $r=\numValidModelR$ for the reduced model and $r=\numValidNullR$ for the estimate. Table~\ref{tab:submission_null} also reports the pooled condition--seed--layer comparisons.

As a null model, the closed-form collapse number of Equation~\eqref{eq:kappa}, evaluated at initialization and mapped through the large-dose asymptote, tracks the same participation minima much less well, including on the settings its derivation covers (Table~\ref{tab:submission_null}). The comparison supports evolving network statistics rather than relying on the initial drive alone.

We retain exits not reached within the integration horizon as censored observations. Exit errors use only uncensored pairs and state their denominator. In the narrow-network condition, the missing model exit means ``not reached by step 1500.'' The gain-$0.1$ condition also has substantial timing error despite the high aggregate correlation. % provenance-ok: condition identity and integration horizon

\paragraph{Numerical resolution.}\label{app:resolution}
To distinguish closure error from numerical error, we repeat three selected predictions with 32, 64 and 128 Gauss--Hermite nodes and with one, two and four Euler substeps per update. Initial states and training horizons remain fixed. Quadrature order has negligible effects here; finer integration shifts exits and participation minima only slightly (Table~\ref{tab:resolution}). These checks support the default resolution in these settings, rather than a convergence guarantee across every condition. % provenance-ok: numerical resolution protocol
% Generated by scripts/review_resolution.py.
\begin{table}[t]\centering\small
\caption{Numerical resolution in three selected settings, each with initialization/feedback seeds $(0,0)$. Exit is in update steps for one, two or four Euler substeps. The last column is the largest absolute change in the first-300-update participation minimum across layers from one to four substeps. These use 64 quadrature nodes; orders 32 and 128 give the same displayed minima and timing. All detailed values and initial states accompany the code.}\label{tab:resolution}
\begin{tabular}{lrrrr}\toprule Setting & Exit: 1 & 2 & 4 & Max. $|\Delta\min p_\ell|$\\\midrule
Baseline & 570 & 567 & 565 & 0.0053\\
Gain 10 & 213 & 209 & 208 & 0.0054\\
Width 100, depth 7 & 2809 & 2801 & 2797 & 0.0023\\
\bottomrule\end{tabular}\end{table}

\begin{figure*}[t]\centering\includegraphics[width=\textwidth]{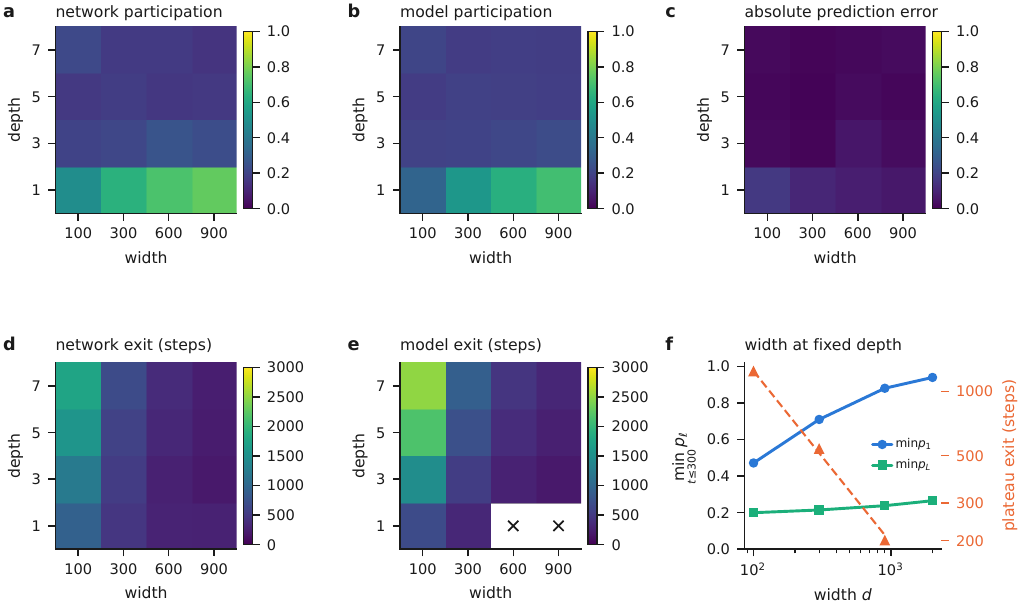}
\caption{\textbf{Width and depth test the reduced model jointly.} (a,b) Deepest-layer participation minima within the first 300 updates in the network and model, averaged over three seeds per cell. (c) Mean absolute error, computed per seed before averaging. (d,e) Plateau exit in update steps, averaged over observed exits; a cross marks a cell with no detected plateau, and an open triangle a cell with a detected plateau but no exit within the horizon. Table~\ref{tab:model_coverage} separates detection failures from censored exits; per-cell counts are saved with the comparison. (f) A separate three-layer sweep with widths 100, 300, 900 and 2000: layer-1 and top-layer participation minima, and plateau exit on the right axis (orange triangles; dashed line: power-law fit).}\label{fig:revision_grid}\end{figure*}

\paragraph{Width and depth.}\label{app:width}
To test width and depth jointly, we use widths $\{100,300,600,900\}$ and depths $\{1,3,5,7\}$, each with three paired seeds and 3,000 steps for both network and model. We initialize each prediction from that network's initial state and retain every grid cell. Across all condition--seed--layer observations, participation has correlation \numGridR\ and mean absolute error \numGridMAE. Figure~\ref{fig:revision_grid}a--c displays the deepest layer cell by cell; panels~d,e compare exit timing. In the separate three-layer width sweep, layer-1 participation rises with width while top-layer participation remains low, and exit becomes earlier (panel~f). The model misses \numGridMissedPlateaus\ detected network plateaus and predicts \numGridExtraPlateaus\ extra ones. Among \numGridExitPairs\ paired exits, median relative error $|t_{\rm model}-t_{\rm network}|/t_{\rm network}$ is \numGridExitError. Three grid settings also occur in the parameter sweeps, so the grid is not wholly independent validation. Coverage, censoring and per-run errors are retained in the analysis outputs (Appendix~\ref{app:code}). % provenance-ok: grid architecture and horizon protocol

 % provenance-ok: width-sweep protocol

\subsection{Alternative recovery approximations}
\label{app:recovery_tests}
Exit-time agreement alone does not establish a particular recovery mechanism. We compare Equation~\eqref{eq:escape-model} with three alternatives on identical per-update collapse trajectories: no accumulated local term ($J_\ell=0$), local growth without the participation factor ($\bar u_\ell$ replaces $\bar u_\ell/\sqrt{p_\ell}$), and transmission initialized from measured hidden-layer label covariances. In the last version, the coefficient $\rho_\ell$ is replaced by
\begin{equation}
 \rho_\ell^{\rm init}=\frac{\|\Lambda_\ell(0)\|_F^2}{\bar u_\ell(0)\|\Lambda_{\ell-1}(0)\|_F^2},
 \qquad \Lambda_0=\Cov(x,y),
\end{equation}
so the initial transmitted amplitudes equal their measured values. These denominators are nonzero in the tested image tasks. No trajectory coefficients are fitted. Each comparison runs to the network's recorded training horizon and uses the original model's first persistent prior-loss band and exit criterion. Network detections retain their smoothed-loss and chance-accuracy conditions.

The full approximation predicts substantially more observed exits than the version without local growth. Removing participation dependence delays many predictions, while substituting measured initial covariances changes them less (Table~\ref{tab:review_recovery}). Covariance trajectories remain approximate: even the baseline's magnitude is not reproduced exactly (Figure~\ref{fig:review_validation}d). This supports the extension as an exit-time approximation, with limited evidence for its full covariance dynamics.

To test configurations outside the original sweeps, we use width 450, depths 2, 4 and 6, and paired initialization--feedback seeds 10, 11 and 12. All four predictions are saved before each 3,000-update training run. These nine runs form the fresh group in Table~\ref{tab:review_recovery}. The earlier parameter and width--depth groups are shown separately because some settings overlap. Coverage retains every run; timing errors are conditional on observed pairs. % provenance-ok: fresh-validation protocol
% Generated by scripts/review_reporting.py.
\begin{table}[t]\centering\scriptsize\setlength{\tabcolsep}{3.5pt}
\caption{Recovery alternatives at each network's recorded training horizon. Both, missed and extra refer to plateau onset detection. Exit pairs count observed model/network exits relative to all observed network exits; timing error is conditional on paired exits. Each variant uses the same collapse trajectory and detector. The fresh group uses new width/depth combinations and seeds, with predictions saved before training.}\label{tab:review_recovery}
\begin{tabular}{lrrrrrr}
\toprule
Group / variant & Runs & Both & Missed & Extra & Exit pairs & Median rel. error\\\midrule
fresh / full & 9 & 9 & 0 & 0 & 9/9 & 0.19\\
fresh / no local & 9 & 9 & 0 & 0 & 0/9 & --\\
fresh / no participation & 9 & 9 & 0 & 0 & 9/9 & 0.96\\
fresh / measured init & 9 & 9 & 0 & 0 & 9/9 & 0.20\\
grid / full & 48 & 39 & 4 & 3 & 39/43 & 0.24\\
grid / no local & 48 & 43 & 0 & 5 & 7/43 & 1.77\\
grid / no participation & 48 & 40 & 3 & 5 & 33/43 & 0.75\\
grid / measured init & 48 & 40 & 3 & 3 & 40/43 & 0.25\\
sweeps / full & 144 & 77 & 1 & 13 & 76/78 & 0.14\\
sweeps / no local & 144 & 78 & 0 & 40 & 4/78 & 0.73\\
sweeps / no participation & 144 & 77 & 1 & 13 & 71/78 & 0.49\\
sweeps / measured init & 144 & 77 & 1 & 13 & 76/78 & 0.14\\
\bottomrule\end{tabular}\end{table}

The scalar closure has an additional analytical boundary: $A_0=0$ and $J_\ell(0)=0$ force every $A_\ell$ and $J_\ell$ to remain zero. It cannot represent recovery driven by nonlinear features on a task with zero population input--label covariance. Such a task could nevertheless be learnable; we do not claim to have tested that possibility here.

Under plain SGD, gain and hidden learning rate combine as $a=\eta_{\rm hid}g$ for fixed normalized feedback and output rate. The recovery law therefore concerns $(a,\eta_{\rm out})$. After merging equivalent rate settings, a joint log--log fit gives exponents \numReviewRateHid\ for $a$ and \numReviewRateOut\ for $\eta_{\rm out}$ ($R^2=\numReviewRateRtwo$), using \numReviewRateRuns\ observed durations among \numReviewRateEligible\ unique effective-rate/seed combinations. Table~\ref{tab:review_rate_fit} gives condition-bootstrap intervals. Gain and hidden-rate sweeps are therefore not independent confirmations. The exponents in Equation~\eqref{eq:recovery} are empirical; their intervals exclude an equal square-root split, while their sum is close to the $-1$ required by joint rate scaling.

% Generated by scripts/review_reporting.py.
\begin{table}[t]\centering\scriptsize\setlength{\tabcolsep}{3.5pt}
\caption{Two-rate recovery fit after merging equivalent SGD gain/rate settings. The fit is conditional on observed plateau durations; intervals resample whole effective-rate conditions (2,000 draws).}\label{tab:review_rate_fit}
\begin{tabular}{lrr}
\toprule
Predictor & Exponent & 95\% interval\\\midrule
$a=\eta_{\rm hid}g$ & -0.40 & [-0.42, -0.38]\\
$\eta_{\rm out}$ & -0.60 & [-0.62, -0.57]\\
\bottomrule\end{tabular}\end{table}

\paragraph{Measurement-window sensitivity.}
The first-300-update participation minimum provides a common reporting window, but not a common dynamical phase across learning rates. We repeat the model comparison with window $T=\min[T_{\rm run},\max(300,300\times10^{-3}/a)]$, where $T_{\rm run}$ is the recorded horizon. Correlation is \numReviewWindowFixedR\ in the fixed-window reintegration and \numReviewWindowRescaledR\ with this rescaled window; mean absolute error changes from \numReviewWindowFixedMAE\ to \numReviewWindowRescaledMAE. All pairs and windows are saved. This is a sensitivity check, not a universal phase detector. % provenance-ok: window protocol

\section{Decodability and gate recovery}
\label{app:representation}
Section~\ref{sec:frozen} distinguishes retained class information from the speed of readout learning. We give the fitting and continuation protocol here, then examine how sensitivity and useful updates change during recovery.

\subsection{Frozen-feature decoding and readout continuation}
\label{app:frozen}
The frozen-feature experiment separates what a representation can support from how quickly the current readout can use it. We replay three baseline initialization--feedback pairs, $(0,0)$, $(1,1)$ and $(2,2)$, with the original minibatches. Saved participation values agree with the original runs to numerical precision. Checkpoints are initialization, each run's deepest-layer participation minimum within the first 300 updates, and step 3000. The geometric checkpoint rule is fixed before evaluating decoders.

For each seed, we sample 10,000 distinct training examples for fitting and 5,000 distinct validation examples for evaluation, using the same examples at every checkpoint. The two sets are disjoint. A least-squares readout fits one-hot targets from centered features with ridge $10^{-3}\operatorname{tr}(\Sigma_h)/d$. Its intercept is the fitting-set target mean; predictions use the largest output coordinate. The ridge is fixed, without selecting it on the evaluation set. Table~\ref{tab:review_features} reports every layer, the covariance statistics on the fitting data, and held-out accuracy. Its normalized covariance statistic uses the fitting examples and the same ridge as the decoder; Appendix~\ref{app:diagnostics} uses a smaller probe and a different regularization level. % provenance-ok: decoder protocol
% Generated by scripts/review_reporting.py.
\begin{table}[t]\centering\scriptsize\setlength{\tabcolsep}{3.5pt}
\caption{Frozen-feature diagnostics with 10,000 training examples for fitting and 5,000 separate validation examples for evaluation. A fixed trace-relative ridge of $10^{-3}$ is used for least-squares one-hot decoding. Here $R$ uses the fitting set and decoder ridge, whereas Table~\ref{tab:E7} uses the probe estimator. The collapse checkpoint is each run's layer-3 participation minimum within its first 300 updates. Values average three seeds.}\label{tab:review_features}
\begin{tabular}{lrrrrr}
\toprule
Checkpoint & Layer & Covariance energy & $R$ & Feature RMS & Decoder accuracy\\\midrule
initial & 1 & $0.472\pm0.012$ & $5.99\pm0.01$ & $0.274\pm0.002$ & $0.877\pm0.007$\\
initial & 2 & $0.372\pm0.029$ & $6.14\pm0.03$ & $0.244\pm0.003$ & $0.890\pm0.006$\\
initial & 3 & $0.300\pm0.016$ & $6.21\pm0.02$ & $0.222\pm0.003$ & $0.893\pm0.007$\\
collapse & 1 & $0.440\pm0.026$ & $6.37\pm0.05$ & $0.227\pm0.003$ & $0.892\pm0.006$\\
collapse & 2 & $0.103\pm0.020$ & $6.31\pm0.06$ & $0.105\pm0.007$ & $0.896\pm0.005$\\
collapse & 3 & $0.014\pm0.005$ & $5.94\pm0.11$ & $0.038\pm0.005$ & $0.886\pm0.005$\\
recovery & 1 & $4.222\pm0.245$ & $7.62\pm0.03$ & $0.498\pm0.013$ & $0.927\pm0.003$\\
recovery & 2 & $7.994\pm0.341$ & $8.21\pm0.04$ & $0.632\pm0.013$ & $0.929\pm0.001$\\
recovery & 3 & $10.565\pm0.287$ & $8.41\pm0.02$ & $0.696\pm0.010$ & $0.924\pm0.002$\\
\bottomrule\end{tabular}\end{table}

To test optimization on the same representation, we freeze the features and continue training the checkpoint's sigmoid readout for 3,000 SGD updates at learning rate $10^{-3}$ and batch size 128. Each control receives identical sampled minibatches. Let $m$ be the fitting-set feature mean and $s=[\E\|h-m\|^2/d]^{1/2}$ its scalar root-mean-square variation. The controls use $h$, $h-m$, or $(h-m)/s$. For the last control we initialize $W'=sW$ and $b'=b+Wm$; for centering alone, $W'=W$ and the same shifted bias. Thus all three begin with exactly the checkpoint's logits. Their subsequent updates differ because ordinary SGD depends on the feature coordinates. Means and scales are estimated only on the fitting set and remain fixed. Figure~\ref{fig:useful_learning}b reports evaluation accuracy; all checkpoints and loss curves are included in the data. These controls establish a difference in learning over a fixed budget. % provenance-ok: continuation protocol

Scalar rescaling has an exact optimizer interpretation. Put $c=h-m$ and write the weights in centered coordinates as $U=W'/s$, so the logit is $Uc+b'$. With fixed $m,s$, plain SGD on $c/s$ gives
\begin{equation}
 \Delta U=-\frac{\eta_{\rm out}}{s^2}\langle e c^\top\rangle,
 \qquad \Delta b'=-\eta_{\rm out}\langle e\rangle.
 \label{eq:rescaled-readout}
\end{equation}
Thus it is equivalent to centered-feature SGD with weight rate $\eta_{\rm out}/s^2$ and unchanged intercept rate, using identical minibatches. This separates an amplitude bottleneck at the chosen optimizer scale from loss of decodability. A single scalar leaves the eigenvalue ratios of the centered feature covariance unchanged; it does not correct anisotropy within that covariance. The code tests the equality of the two update trajectories.

\paragraph{Covariance amplitude and normalization.}
Table~\ref{tab:E7} compares raw and normalized label covariance at the same checkpoint, using the probe statistic defined in Appendix~\ref{app:diagnostics}. The marked amplitude decrease in baseline DFA coexists with a much smaller change in normalized covariance. This motivates the separate held-out decoding test; the probe statistic itself is not an accuracy estimate.
% Generated by scripts/supplement_tables.py.
\begin{table}[t]\centering\footnotesize\setlength{\tabcolsep}{3.5pt}
\caption{Covariance amplitude and its normalized statistic at matched layer-3 checkpoints. Within each run, $t_*\ge30$ minimizes covariance energy among checkpoints with a recorded $R_3$. The energy ratio is $\|\Lambda_3(0)\|_F^2/\|\Lambda_3(t_*)\|_F^2$. A large amplitude loss need not imply a comparable loss of normalized covariance. $R_3$ is not held-out accuracy.}\label{tab:E7}
\begin{tabular}{lrrrrr}\toprule
Condition & $n$ & $t_*$ & Energy ratio & $R_3(0)$ & $R_3(t_*)$\\\midrule
Plain DFA & 3 & $80\pm0$ & $28.0\pm6.6$ & $7.64\pm0.03$ & $7.42\pm0.10$\\
Backpropagation & 3 & $147\pm12$ & $1.2\pm0.0$ & $7.64\pm0.03$ & $7.74\pm0.00$\\
Pre-aligned DFA & 3 & $40\pm0$ & $0.9\pm0.0$ & $7.64\pm0.03$ & $7.86\pm0.01$\\
Center teaching signal & 3 & $40\pm0$ & $0.8\pm0.0$ & $7.64\pm0.03$ & $7.69\pm0.01$\\
Frozen readout & 3 & $253\pm50$ & $7785.5\pm5851.3$ & $7.64\pm0.03$ & $5.63\pm0.26$\\
Feedback gain 3 & 3 & $40\pm0$ & $187.4\pm56.5$ & $7.64\pm0.03$ & $7.01\pm0.11$\\
Balanced softmax & 3 & $40\pm0$ & $0.6\pm0.0$ & $7.63\pm0.03$ & $7.83\pm0.01$\\
Imbalanced softmax & 3 & $40\pm0$ & $1.6\pm0.7$ & $8.06\pm0.16$ & $8.96\pm0.08$\\
\bottomrule\end{tabular}\end{table}

\begin{figure*}[t]\centering\includegraphics[width=\textwidth]{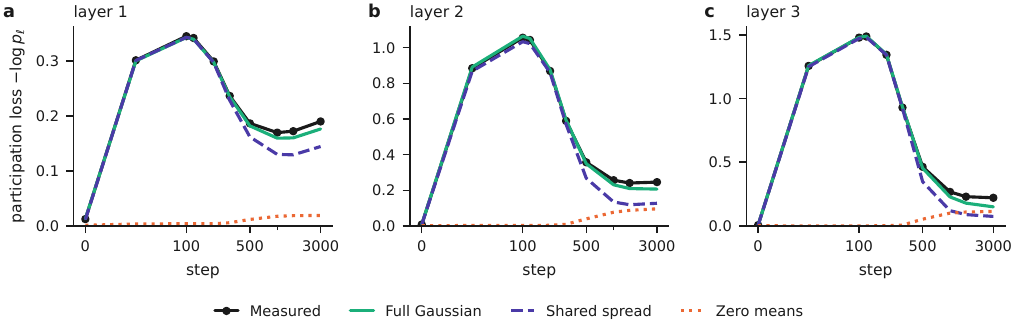}
\caption{\textbf{Unit means and spreads in the Gaussian reconstruction.} Panels (a--c) show layers 1--3. Measured participation loss is compared with the full per-unit Gaussian reconstruction and two moment substitutions: one shared root-mean-square spread with measured unit means, or zero means with measured unit spreads. Curves average $-\log p_\ell$ across \numGaussRuns\ baseline trajectories at each saved snapshot. The horizontal axis is linear through step 100 and logarithmic thereafter. These substitutions change the reconstruction, not the network's training.}\label{fig:gaussian_ablations}\end{figure*}

\subsection{Reconstructing participation from unit means and spreads}
\label{app:gate_moments}
The full Gaussian reconstruction uses both the location and spread of each unit's preactivation distribution. To assess their contributions to the early participation decline, we repeat the reconstruction defined in Appendix~\ref{app:diagnostics} with two substitutions. The \emph{shared-spread} reconstruction retains every measured $\mu_{\ell i}$ but replaces each $\sigma_{\ell i}$ by $s_\ell=(d_\ell^{-1}\sum_j\sigma_{\ell j}^2)^{1/2}$, computed separately for each run and snapshot. The \emph{zero-mean} reconstruction sets every $\mu_{\ell i}=0$ while retaining the measured $\sigma_{\ell i}$. Both recompute gate energies and then participation using Equation~\eqref{eq:participation}.

Figure~\ref{fig:gaussian_ablations}a--c compares these reconstructions with measured participation loss, $-\log p_\ell$, in all three layers. The full reconstruction matches this loss closely during early collapse (median reconstructed-to-measured ratio \numGaussRatioEarly\ over steps 50--300), but departs later. At step 100, the shared-spread reconstruction gives a median reconstructed-to-measured loss ratio of \numGaussSharedHundred\ across runs and layers; the zero-mean reconstruction gives only \numGaussZeroMeanHundred. Thus, differences between unit means suffice to reproduce the early concentration when a common within-unit spread is retained. Differences in spread alone produce little concentration at that time. Later, both sets of moments matter: at step 3000 the ratios are \numGaussSharedLate\ and \numGaussZeroMeanLate, compared with \numGaussRatioLate\ for the full reconstruction.

These are conditional comparisons within the Gaussian approximation. They do not partition a causal effect, because the moments are measured after training and participation depends nonlinearly on both. They also do not remove the late mismatch of the full reconstruction.

\subsection{Gate turnover and useful descent}
\label{app:gates}\label{app:gate_diagnostics}
\begin{figure*}[t]\centering\includegraphics[width=\textwidth]{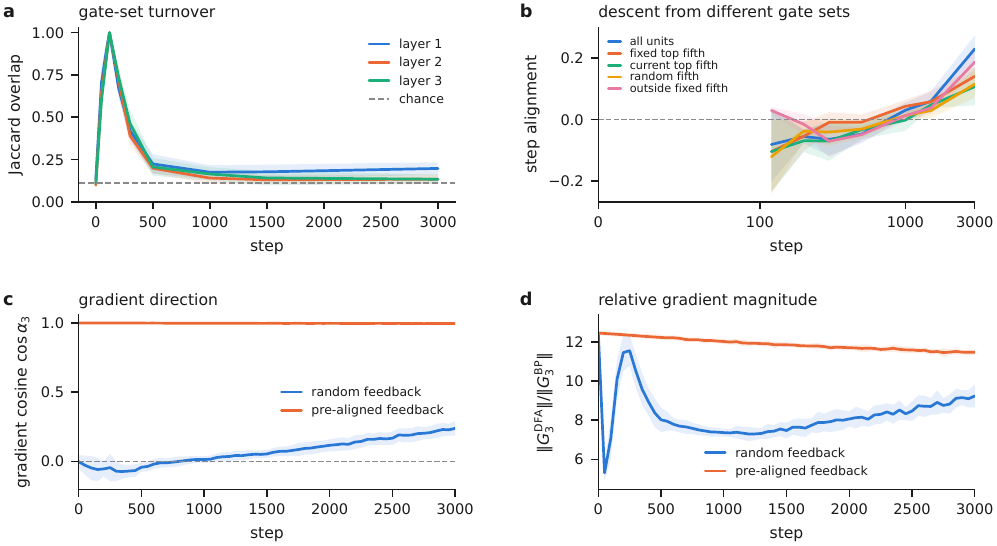}
\caption{\textbf{Gate turnover and the direction and size of learning signals.} (a) Jaccard overlap of the current top-fifth gate-energy set with its step-120 set; dashed line: chance overlap. (b) Layer-3 alignment of norm-matched virtual weight-and-bias updates from all units, current or fixed top-fifth sets, a random fifth, and the fixed set's complement; time is linear through step 100 and logarithmic thereafter. (c,d) Layer-3 raw weight-gradient cosine and DFA-to-BP norm ratio for random and pre-aligned feedback. Curves and bands show means and one standard deviation: 15 trajectories in a,c,d and three paired seeds in b. The complement in b contains more units, but every virtual step has the same norm.}\label{fig:gates}\end{figure*}

Participation rebounds as learning resumes (Figure~\ref{fig:phenomenon}f), but the identities of sensitive units change. Figure~\ref{fig:gates}a follows the top-fifth gate-energy set selected at step 120. Its overlap with the currently selected set falls toward the chance reference, arguing against recovery through one persistent selected channel. This does not exclude predictive value in early gate statistics.

 % provenance-ok: gate-set comparison counts

\paragraph{Which units support descent?}\label{app:masked}
Turnover alone does not identify the units responsible for learning. Norm-matched, single-layer virtual updates test the direction of descent independently of their raw magnitude (Appendix~\ref{app:diagnostics}). The fixed early set retains useful direction, but its complement also supports descent later (Figure~\ref{fig:gates}b). These snapshot tests do not establish recovery after permanent ablation.

\paragraph{Direction and magnitude.}
The large projected descent under pre-aligned feedback reflects both a nearly BP-aligned direction and a large gradient-norm ratio (panels~c,d). Gate participation determines neither factor. Figure~\ref{fig:useful_learning}c shows their distinction from sensitivity during the baseline trajectory; Table~\ref{tab:diagnostic_taxonomy} summarizes the different measurements.

\section{Interventions, generality and boundary cases}
\label{app:controls}
The controls in this section ask when the mechanism matters and when interventions based on it fail. We compare intervention costs and instabilities, then test changes in optimizer, architecture, readout and data. Complete condition summaries accompany the code.

\begin{figure*}[t]\centering\includegraphics[width=\textwidth]{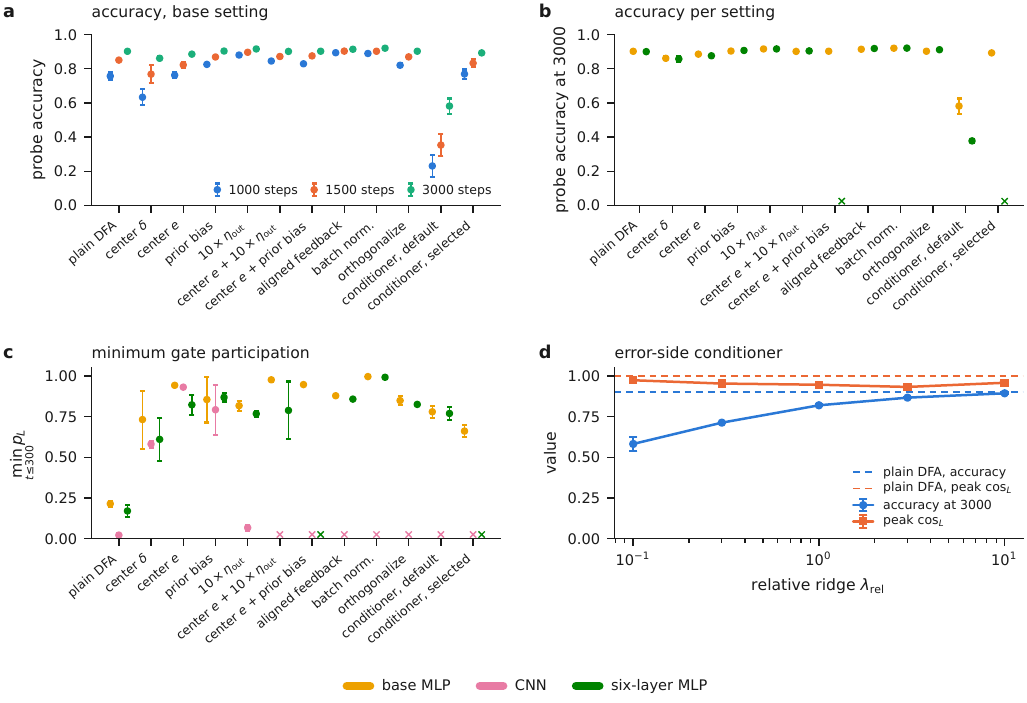}
\caption{\textbf{Interventions.} All networks, including the CNN, are trained on MNIST. (a) Probe accuracy at 1{,}000, 1{,}500 and 3{,}000 steps on the base network. (b) Accuracy at 3{,}000 steps in the base and six-layer MLPs; CNN runs end earlier. (c) Minimum top-layer participation within 300 steps in both MLPs and the CNN. Points and bars show means and one standard deviation; crosses mark unmeasured conditions. (d) Accuracy at step 3000 and peak cosine versus relative ridge $\lambda_{\rm rel}$; dashed lines show plain DFA. The conditioner appears at its default ridge $0.1$ and, for the base MLP, at the ridge $10$ selected by the same step-3000 validation accuracy. The \emph{orthogonalize} control uses Muon-style orthogonalization, not full Muon. All measured conditions use three paired seeds.}\label{fig:cures}\end{figure*} % provenance-ok: conditioner settings and endpoints

\subsection{Intervention comparisons}
\label{app:intervention_comparisons}
Preventing collapse need not improve accuracy at every training time, because an intervention can also slow readout adaptation. Error centering lowers top-layer mean-activity energy at step 300 from \numBaseHLThree\ to \numCenterEHL. At step 1500, mean-error norm remains \numCenterEEbarFifteen\ and loss \numCenterELossFifteen\ (baseline: \numBaseEbarFifteen\ and \numBaseLossFifteen). Thus, its accuracy cost depends on when performance is measured (Figure~\ref{fig:cures}a), even when early gate concentration is reduced. Comparing accuracy with participation across architectures makes the distinction explicit (panels~b,c).

The local-error conditioner offers another trade-off (Figure~\ref{fig:cures}d). Relative ridge is selected from $\{0.1,0.3,1,3,10\}$ by mean validation accuracy at step 3000, using the same three seed pairs reported in the figure. The default and selected settings are shown separately; selection is descriptive because there is no independent evaluation set. The selected ridge retains high cosine. A plateau is detected in only one run at the selected ridge, so Table~\ref{tab:review_timing} compares a common sustained-loss learning criterion across all runs. The conditioner also costs more than centering (Table~\ref{tab:cures}). % provenance-ok: ridge selection protocol

Table~\ref{tab:E10} reports the large finite losses seen with some non-saturating activations; these should not be equated with an early stop from nonfinite loss. Because a linear unit's gate is constant, the gate--error residual cannot explain its failure. Even with zero mean teaching signal, local covariance can shift unit means through nonzero presynaptic means (Appendix~\ref{app:drift}). The logs show large finite growth consistent with this route, rather than establishing its contribution to the full trajectory. With raw pixels, top-layer mean-activity energy grows from \numInstabLinearHzero\ to $\numInstabLinearHend$ in the linear network and reaches $\numInstabReluHend$ with ReLU and $\numInstabGeluHend$ with GELU. With per-pixel input centering at the larger rate $10^{-3}$ rather than $3\times10^{-4}$, mean activity stays small in the linear network (mean per-run peak energy \numInstabLinearCenteredHmax). ReLU and GELU retain larger activity means. These controls do not isolate covariance-driven motion from upstream changes or residual terms, and finite-horizon growth does not establish mathematical divergence. % provenance-ok: learning rates
% Generated by scripts/supplement_tables.py.
\begin{table}[t]\centering\footnotesize\setlength{\tabcolsep}{3.5pt}
\caption{Final loss with non-saturating units after 3000 updates. All 27 activation--protocol combinations are retained, with three seeds each. Entries give the mean (standard deviation) in scientific notation. The hidden learning rate is shown; a faster readout uses ten times that rate. Large finite losses indicate instability and are not nonfinite early stops. The constant-predictor reference is about $\numPriorLoss$.}\label{tab:E10}
\begin{tabular}{lrrrr}\toprule
Protocol & $\eta_{\rm hid}$ & ReLU loss & GELU loss & Linear loss\\\midrule
Plain DFA & $10^{-4}$ & $2.9(0.3)\!\times10^{0}$ & $2.8(0.15)\!\times10^{0}$ & $1.8(0.083)\!\times10^{0}$\\
Center error & $10^{-4}$ & $3.2(0.23)\!\times10^{0}$ & $3.7(0.15)\!\times10^{0}$ & $1.4(1.3)\!\times10^{1}$\\
Plain DFA & $3\times10^{-4}$ & $1(0.027)\!\times10^{0}$ & $1(0.041)\!\times10^{0}$ & $9.8(0.3)\!\times10^{-1}$\\
Center error & $3\times10^{-4}$ & $1.5(1.3)\!\times10^{2}$ & $1.4(1.2)\!\times10^{3}$ & $7.4(9.3)\!\times10^{10}$\\
Center error + faster readout & $3\times10^{-4}$ & $2.3(2)\!\times10^{2}$ & $2.4(1.9)\!\times10^{3}$ & $1.3(1)\!\times10^{5}$\\
Backpropagation & $10^{-3}$ & $1.4(0.058)\!\times10^{0}$ & $1.7(0.091)\!\times10^{0}$ & $9.9(0.19)\!\times10^{-1}$\\
Prior-bias initialization & $10^{-3}$ & $5.8(0.19)\!\times10^{-1}$ & $5.7(0.16)\!\times10^{-1}$ & $7.2(0.13)\!\times10^{-1}$\\
Center error + faster readout & $10^{-3}$ & $4.6(4.7)\!\times10^{9}$ & $5.3(4.5)\!\times10^{9}$ & $5.1(3.9)\!\times10^{12}$\\
Center error + inputs & $10^{-3}$ & $5.5(0.056)\!\times10^{-1}$ & $1.3(1.2)\!\times10^{0}$ & $3.1(0.011)\!\times10^{0}$\\
\bottomrule\end{tabular}\end{table}

\begin{figure*}[t]\centering\includegraphics[width=\textwidth]{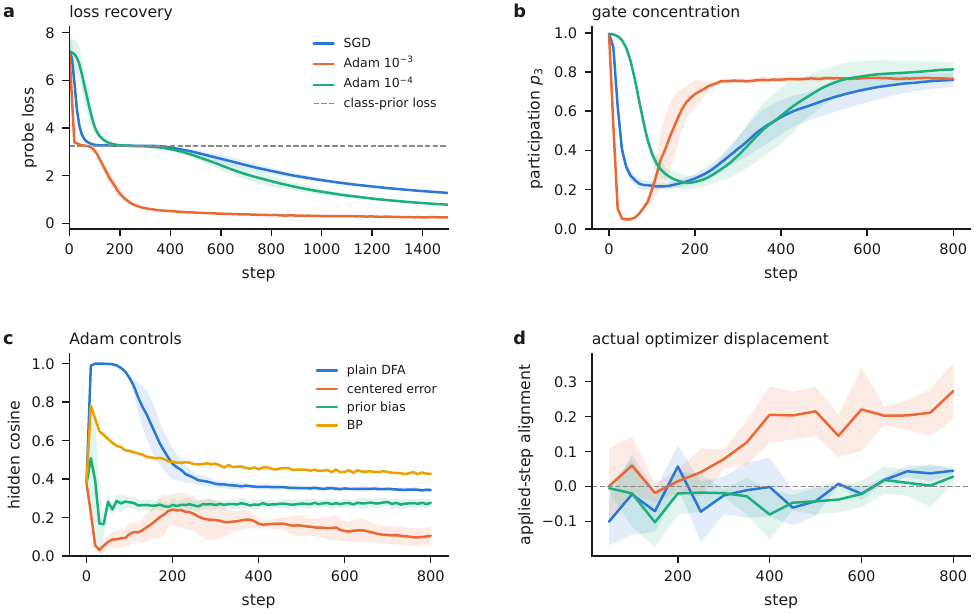}
\caption{\textbf{Adam produces deeper gate concentration with faster loss recovery.} (a) Probe loss and the prior-loss reference. (b) Layer-3 participation. (c) Mean-error controls and BP at the larger Adam learning rate. (d) Layer-3 cosine between the actual displacement and negative true gradient, including hidden biases. Colours in b,d follow a; panel c has its own legend. Curves show means and bands show one standard deviation across three seed pairs. The SGD dose formula is not applied to Adam.}\label{fig:revision_adam}\end{figure*}

\subsection{Optimizer dependence}
\label{app:optimizer}
To test optimizer dependence, we compare Adam \citep{kingma2015} and SGD at matched initializations, feedback draws, minibatches and update counts. Adam uses learning rates $10^{-3}$ and $10^{-4}$, $\beta_1=0.9$, $\beta_2=0.999$ and $\epsilon=10^{-8}$, where $\beta_1,\beta_2$ are the moving-average decay factors and $\epsilon$ stabilizes the denominator; weight decay is zero. Each rate is tested with plain DFA, centered error, prior-bias initialization and BP, using three seed pairs and 3,000 updates. The matched SGD baselines also use three seeds; this is not a hyperparameter search. % provenance-ok: optimizer protocol

The larger Adam rate produces deeper gate concentration but faster learning (Figure~\ref{fig:revision_adam}a,b). The plateau detector identifies a plateau in only one of three runs, so its exit at \numAdamFastExit\ is not a three-run mean. We instead compare the first logged loss below $0.9\mathcal L_{\rm prior}$ that remains below it for at least 50 updates, without requiring a preceding plateau. Every Adam and SGD baseline run reaches this criterion: means are \numReviewAdamLearn\ for larger-rate Adam, \numReviewAdamSlowLearn\ for smaller-rate Adam, and \numReviewSGDLearn\ for SGD (Table~\ref{tab:review_timing}). Minimum participation is \numAdamFastP\ for larger-rate Adam versus \numAdamSGDP\ for SGD. Centering and prior-bias initialization suppress the high-cosine transient (panel~c). The complete control summary retains all detected exits with their denominators. % provenance-ok: learning-time criterion
% Generated by scripts/review_reporting.py.
\begin{table}[t]\centering\scriptsize\setlength{\tabcolsep}{3.5pt}
\caption{Learning time without requiring a detected plateau: first probe loss below 90\% of the constant-predictor loss, remaining below it for 50 updates. All runs use the same criterion and seed pairs $(0,0)$, $(1,1)$ and $(2,2)$; the shared SGD baseline is shown once, so its accuracy can differ slightly from the 15-trajectory value in Table~\ref{tab:E1}. Values are means $\pm$ standard deviations; counts state how many reach it within 3,000 updates.}\label{tab:review_timing}
\begin{tabular}{lrrrr}
\toprule
Condition & $n$ & Time & Reached/$n$ & Final accuracy\\\midrule
Plain SGD & 3 & $537\pm47$ & 3/3 & $0.902\pm0.005$\\
Adam, $\eta=10^{-3}$ & 3 & $117\pm6$ & 3/3 & $0.969\pm0.004$\\
Adam, $\eta=10^{-4}$ & 3 & $480\pm36$ & 3/3 & $0.928\pm0.006$\\
Conditioner, $\lambda_{\rm rel}=0.1$ & 3 & $1897\pm67$ & 3/3 & $0.582\pm0.044$\\
Conditioner, $\lambda_{\rm rel}=3$ & 3 & $433\pm21$ & 3/3 & $0.866\pm0.005$\\
Conditioner, $\lambda_{\rm rel}=10$ & 3 & $387\pm23$ & 3/3 & $0.893\pm0.007$\\
\bottomrule\end{tabular}\end{table}

An adaptive optimizer changes the direction and size of the raw DFA update. To measure its effect on useful descent (Figure~\ref{fig:revision_adam}d), let $\theta_\ell$ collect hidden layer $\ell$'s weights and biases. We record its actual displacement $\Delta\theta_\ell$, norm and first-order loss decrease $-(\nabla_{\theta_\ell}\mathcal L)^\top\Delta\theta_\ell$ on the current minibatch. Directional alignment is $\cos(\Delta\theta_\ell,-\nabla_{\theta_\ell}\mathcal L)$, so positive values favour descent. The larger-rate Adam update becomes positively aligned during early recovery, while SGD and smaller-rate Adam remain near zero or negative for longer (Figure~\ref{fig:revision_adam}d). These diagnostics include momentum and coordinatewise scaling. They are distinct from the raw weight-gradient projection in Equation~\eqref{eq:projection}.

\subsection{Reconstructing the original DFA protocol}
\label{app:nokland}
Our baseline shares the readout, loss, hidden nonlinearity and input scaling of \citet{nokland2016} but uses plain SGD and Xavier initialization. To reconstruct that setting, we train the original $3\times800$ tanh MNIST network with logistic outputs and binary cross-entropy, inputs in $[0,1]$, RMSprop without momentum or weight decay, and minibatches of 64. DFA starts from zero weights and biases with feedback entries drawn uniformly from $[-1/\sqrt{d_\ell},1/\sqrt{d_\ell}]$; BP weights and biases are drawn uniformly from $[-1/\sqrt{n},1/\sqrt{n}]$ for fan-in $n$. A third arm keeps our Xavier initialization and Gaussian feedback under the same optimizer. The original learning rate, decay and $\epsilon$ are not reported; we use the PyTorch implementation with decay $0.99$ and $\epsilon=10^{-8}$, which, like the original RMSprop, has no bias correction, and test rates $10^{-3}$ and $10^{-4}$, each with three seed pairs and 6,000 updates, the example count of the baseline (Table~\ref{tab:submission_nokland}). % provenance-ok: original protocol constants

At zero initialization, hidden layers beyond the first have zero weight gradients and raw bias gradient $G_{b,\ell}=B_\ell\ebar$, a purely common-mode signal. RMSprop rescales this gradient using an exponential average of its square. With zero-initialized second moments and negligible $\epsilon$, the first displacement of a bias coordinate $i$ with nonzero gradient is
\[
\Delta b_{\ell i}\simeq-\frac{\eta}{\sqrt{1-0.99}}\,\operatorname{sign}((B_\ell\ebar)_i) % provenance-ok: RMSprop decay
=-10\eta\,\operatorname{sign}((B_\ell\ebar)_i).
\] % provenance-ok: RMSprop decay and first-step factor
For a constant gradient, the displacement approaches magnitude $\eta$. More generally, the initial coordinate scaling permits a local change in mean preactivation of up to $10\eta(\|\hb_{\ell-1}\|_1+1)$, where $\|\cdot\|_1$ sums absolute entries. When the mean term determines the gradient signs, this shift can be large even for small feedback scale or $\|\ebar\|$.

At the larger rate, at least 99.9\% of top-layer units saturate within ten updates in all three arms, including BP. DFA leaves chance accuracy after \numNokDFAZeroFastChance\ updates, whereas BP stays at chance longer (\numNokBPFaninFastChance\ updates on average, varying widely across seeds). A plausible reason is that BP's error reaches a hidden layer only through every saturated layer above it, whereas DFA delivers error to each layer through a single derivative gate. At the smaller rate, few top-layer units pass $|h|>0.9$, and DFA leaves chance after \numNokDFAZeroSlowChance\ updates, against \numNokBPFaninSlowChance\ for BP. In fully saturated runs, gate participation stays high because all gate energies are equally small (Appendix~\ref{app:diagnostics}); the saturated fraction is the informative measure. % provenance-ok: saturation threshold and fraction reported in the table
% Generated by scripts/submission_reporting.py.
\begin{table}[t]\centering\scriptsize\setlength{\tabcolsep}{3pt}
\caption{Reconstruction of the original DFA protocol \citep{nokland2016}: $3\times800$ tanh MLP, logistic outputs with binary cross-entropy, inputs in $[0,1]$, RMSprop and minibatches of 64, with zero initialization and uniform feedback for DFA and fan-in initialization for BP; the last rows keep Xavier initialization and Gaussian feedback. Saturated is the largest fraction of top-layer units with $|\bar h_{3i}|>0.9$. Leaves chance is the first update with probe accuracy above the majority-class rate plus 0.05; learning time uses the criterion of Table~\ref{tab:review_timing}; the 80\% column is the first update reaching that accuracy; the last column is accuracy after 6,000 updates. At zero initialization, layers above the first are input-independent until their weights grow, so their peak cosine is one by construction. Means $\pm$ standard deviations over three seed pairs.}\label{tab:submission_nokland}
\begin{tabular}{llrrrrrrr}
\toprule
Protocol & Rate & $n$ & Peak $\cos_3$ & Saturated & Leaves chance & Learning time & 80\% acc. & Accuracy\\\midrule
DFA, zero initialization & $10^{-3}$ & 3 & $1.000\pm0.000$ & $1.00\pm0.00$ & $40\pm0$ & $40\pm0$ & $60\pm0$ & $0.973\pm0.005$\\
BP, fan-in initialization & $10^{-3}$ & 3 & $1.000\pm0.000$ & $1.00\pm0.00$ & $223\pm162$ & $273\pm189$ & $467\pm151$ & $0.974\pm0.004$\\
DFA, Xavier initialization & $10^{-3}$ & 3 & $1.000\pm0.000$ & $1.00\pm0.00$ & $40\pm0$ & $40\pm0$ & $60\pm0$ & $0.972\pm0.010$\\
DFA, zero initialization & $10^{-4}$ & 3 & $1.000\pm0.000$ & $0.12\pm0.01$ & $47\pm6$ & $70\pm0$ & $187\pm6$ & $0.959\pm0.002$\\
BP, fan-in initialization & $10^{-4}$ & 3 & $0.954\pm0.002$ & $0.25\pm0.00$ & $10\pm0$ & $10\pm0$ & $43\pm6$ & $0.969\pm0.003$\\
DFA, Xavier initialization & $10^{-4}$ & 3 & $0.987\pm0.004$ & $0.29\pm0.12$ & $40\pm0$ & $80\pm10$ & $220\pm10$ & $0.967\pm0.001$\\
\bottomrule\end{tabular}\end{table}

The saturation shared by BP and DFA at the larger rate is consistent with RMSprop's large initial normalized steps. With Adam at the same rate (Appendix~\ref{app:optimizer}), BP does not exhibit the same near-unit-cosine transient (peak top-layer cosine \numAdamBPCos, minimum participation \numAdamBPP), whereas DFA shows strong gate concentration (minimum participation \numAdamFastP). The Adam comparison uses the baseline architecture, initialization and batch size, so this contrast does not isolate optimizer bias correction. Both adaptive-optimizer experiments shorten the DFA transient relative to the hundreds of updates observed with plain SGD in the main experiments.

\subsection{Sign-error feedback}
\label{app:sign}
Sign-error DFA broadcasts $\operatorname{sign}(e)$ instead of $e$ \citep{frenkel2021}. With independent sigmoid outputs and binary targets, $0<\hat y_c<1$ gives $\operatorname{sign}(e_c)=1-2y_c$. Before derivative gating, the broadcast is therefore a fixed target projection plus the constant $B_\ell\mathbf1$. The population mean of the sign vector is $\mathbf1-2\pi$, with norm $\sqrt C\,|1-2/C|$ for balanced one-hot targets. Batch means fluctuate around this fixed vector instead of shrinking as the readout learns the prior. Derivative gates still change during training, so the gated teaching signal is not fixed. The readout keeps its true gradient. We train the MNIST baseline with this signal, with prior-bias initialization, and with the batch mean of $\operatorname{sign}(e)$ subtracted before projection, using three seed pairs and 3,000 updates (Table~\ref{tab:submission_sign}). % provenance-ok: training horizon

Because $\operatorname{sign}(e)$ does not depend on the readout, prior-bias initialization leaves every hidden update unchanged, and only readout-driven differences in accuracy remain. The plain signal collapses the top layer within the first updates and keeps it collapsed through training, with low participation and slow learning. Subtracting the batch mean prevents sustained collapse and improves learning to the level of error-centered DFA. Both versions use the same discrete sign values before centering. The comparison therefore supports persistent mean drive as a cause of the sustained collapse, rather than quantization alone.
% Generated by scripts/submission_reporting.py.
\begin{table}[t]\centering\scriptsize\setlength{\tabcolsep}{3.5pt}
\caption{Sign-error feedback in the MNIST baseline: hidden layers receive $\operatorname{sign}(e)$ through the baseline feedback, and the readout keeps its gradient. Participation minima use the first 300 updates; final values and accuracy are at step 3000. Learning time uses the criterion of Table~\ref{tab:review_timing}. Plain sign feedback and prior-bias initialization give identical hidden updates, because $\operatorname{sign}(e)=\mathbf 1-2y$ does not depend on the readout. Means $\pm$ standard deviations over three seed pairs.}\label{tab:submission_sign}
\begin{tabular}{lrrrrrrr}
\toprule
Condition & $n$ & Peak $\cos_3$ & Final $\cos_3$ & Min.\ $p_3$ & Final $p_3$ & Learning time & Accuracy\\\midrule
Sign & 3 & $1.000\pm0.000$ & $0.957\pm0.010$ & $0.013\pm0.003$ & $0.055\pm0.008$ & $1833\pm398$ & $0.750\pm0.031$\\
Sign, prior bias & 3 & $1.000\pm0.000$ & $0.957\pm0.010$ & $0.013\pm0.003$ & $0.055\pm0.008$ & $1830\pm393$ & $0.750\pm0.031$\\
Sign, centered & 3 & $0.378\pm0.021$ & $0.026\pm0.001$ & $0.892\pm0.021$ & $0.872\pm0.009$ & $1073\pm31$ & $0.896\pm0.005$\\
\bottomrule\end{tabular}\end{table}

\begin{figure*}[t]\centering\includegraphics[width=\textwidth]{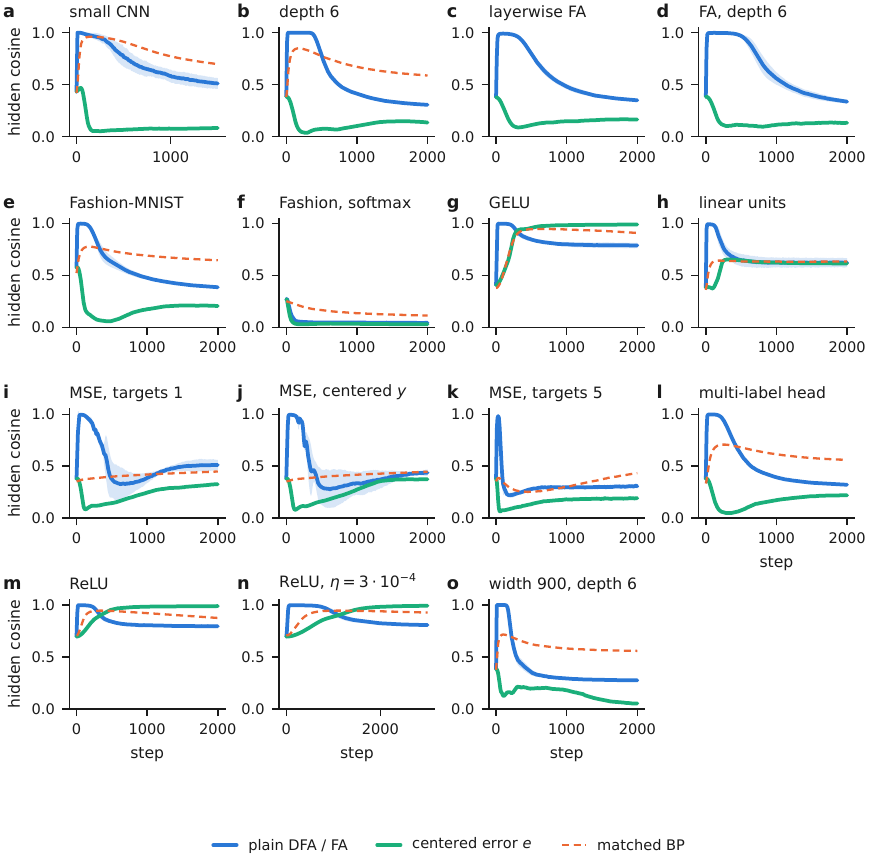}
\caption{\textbf{Generality across protocols.} Deepest-layer cosine under plain DFA (layerwise FA in c,d), centered error, and matched BP where available; each BP curve is a single run. Panel titles specify the protocols; MSE denotes squared error. Curves show run means; bands show one standard deviation for plain DFA/FA. ReLU and GELU cosines require complementary gate and loss measurements.}\label{fig:generality}\end{figure*}

\subsection{Changes in architecture, activation and objective}
\label{app:generality}
Architecture and feedback routing do not confine the transient to a shallow DFA network. The CNN, six-layer MLP and layerwise FA networks show strong activity convergence (Figure~\ref{fig:generality}a--d), which persists at greater width (panel~o). In the width-300 six-layer tanh MLP (panel~b), layers 4--6 reach cosine \numDepthSixCos\ and the plateau lasts \numDepthSixDuration\ steps. The CNN's first block reaches cosine \numCNNConvOneCos\ and its deepest block minimum participation \numCNNDeepP\ (one matched BP run: \numCNNConvOneBP\ and \numCNNDeepBPP). % provenance-ok: six-layer architecture

The Fashion-MNIST comparison shows the effect of sigmoid versus softmax readouts on the same dataset (panels~e,f). Squared-error controls separate target centering from prediction centering and show how target scale changes the relative input-dependent error (panels~i--k). The augmented twelve-output classification target also exhibits a transient (panel~l). Large-target squared error gives a mean-to-input-dependent error ratio of \numMSERatio, compared with \numBCERatio\ for baseline sigmoid outputs. Across from-scratch runs, this ratio at step 10 correlates with peak cosine at $r=\numGenRCos$ (95\% CI \numGenRCosCI; \numGenN\ trajectories across \numGenConditions\ settings). The percentile interval resamples whole physical settings with all their runs retained, using 2,000 draws; it describes uncertainty across settings rather than independent run-level replication. % provenance-ok: diagnostic step and confidence level

GELU, linear and ReLU units lack tanh's two-sided saturation (panels~g,h,m,n). Positive activity bias can raise cosine without suppressing input dependence; GELU is not strictly nonnegative. Gate and loss measurements are therefore needed. Error centering can also destabilize these networks, with the ReLU comparison showing learning-rate dependence (panels~m,n).

\paragraph{Mean predictions with centered targets.}
A random readout acting on nonzero mean hidden activity can produce nonzero mean predictions, and hence a drive even with centered targets. In the squared-error control, centered targets still give $\|\ebar_0\|=\numHeadMSECtrEbar$, dose \numHeadMSECtrKappa\ and peak cosine \numHeadMSECtrCos. Zeroing the output weights reduces the initial error to \numHeadMSECtrWzeroEbar\ and the cosine to \numHeadMSECtrWzeroCos. The same intervention does not rescue sigmoid outputs, whose zero-logit prediction is one half at initialization (cosine \numHeadSigmoidWzeroCos; Table~\ref{tab:E9}).
% Generated by scripts/supplement_tables.py.
\begin{table}[t]\centering\footnotesize\setlength{\tabcolsep}{3.5pt}
\caption{Readout initialization and target centering in the baseline architecture. Zero readout means zero output weights; sigmoid predictions are then one half at initialization. Peak cosine uses the 2000-update trajectory, and participation minima use the first 300 updates.}\label{tab:E9}
\begin{tabular}{lrrrr}\toprule
Readout and targets & $n$ & Initial $\|\bar e\|$ & Peak cosine & Min. $p_3$\\\midrule
Centered squared error, random readout & 3 & $0.803\pm0.346$ & $0.995\pm0.002$ & $0.366\pm0.066$\\
Centered squared error, zero readout & 3 & $0.022\pm0.006$ & $0.378\pm0.020$ & $0.987\pm0.001$\\
Same zero readout, BP & 3 & $0.022\pm0.006$ & $0.378\pm0.021$ & $0.994\pm0.001$\\
Uncentered squared error, zero readout & 3 & $0.317\pm0.000$ & $0.752\pm0.014$ & $0.944\pm0.007$\\
Sigmoid, zero readout & 3 & $1.265\pm0.000$ & $0.998\pm0.000$ & $0.243\pm0.014$\\
Softmax, zero readout & 3 & $0.022\pm0.006$ & $0.250\pm0.007$ & $0.940\pm0.010$\\
\bottomrule\end{tabular}\end{table}

\begin{figure*}[t]\centering\includegraphics[width=\textwidth]{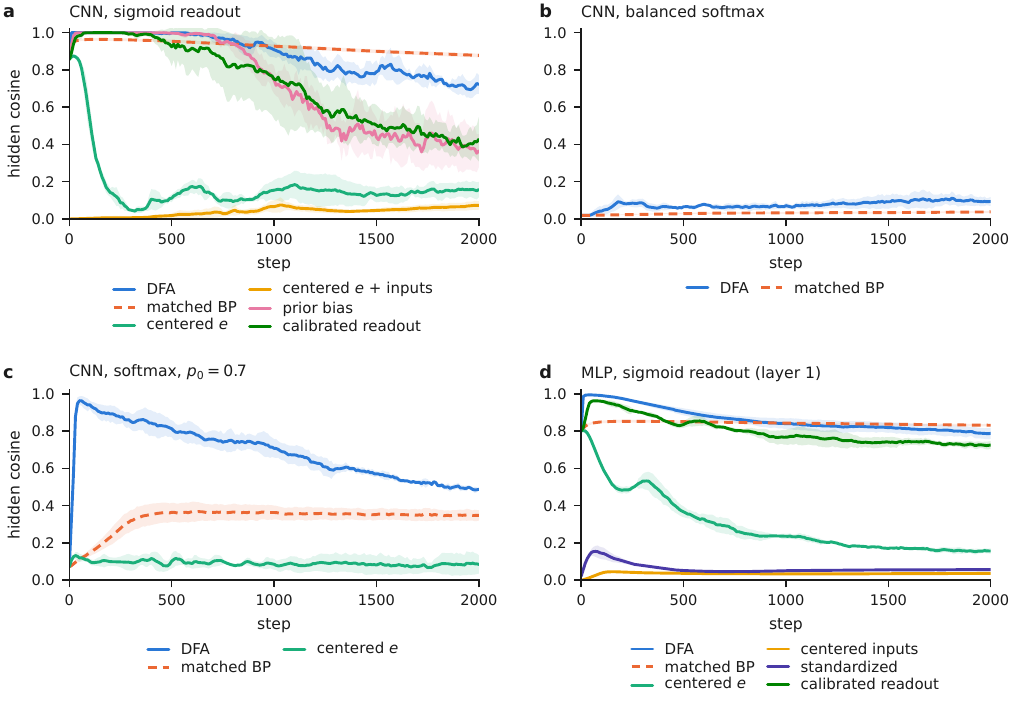}
\caption{\textbf{CIFAR-10.} Hidden cosine of CNN block 3 (the 128-unit fully connected hidden layer; a--c) or MLP layer 1 (d) under DFA, the matched BP network and the interventions. (a) Small CNN with sigmoid readout, comparing prior-bias initialization, mean-logit calibration and centering controls. (b) CNN with balanced softmax and standardized inputs. (c) CNN with softmax and class-0 mixture weight $p_0=0.7$. (d) Flattened-input MLP with error centering, per-pixel centering, global standardization or mean-logit calibration. BP uses matched initialization and minibatches. Curves show means across three seed pairs, bands one standard deviation. High cosine alone does not establish collapse; Table~\ref{tab:cifar_controls} reports complementary measurements.}\label{fig:cifar}\end{figure*} % provenance-ok: sampling-mixture protocol

\subsection{CIFAR-10 controls and a learned MNIST prior shift}
The larger input mean on CIFAR-10 tests whether reducing mean output error is sufficient to prevent collapse. It is not sufficient in the tanh CNN with a sigmoid readout: prior-bias initialization and mean-logit calibration leave substantial activity convergence, whereas combining input and error centering reduces it (Figure~\ref{fig:cifar}a). With standardized inputs, balanced softmax largely avoids this transient, but an imbalanced prior restores it (panels~b,c). Balanced-softmax benchmarks \citep{sanfiz2021} therefore resemble our low-dose control more than the uncalibrated sigmoid baseline.

The flattened MLP makes the role of the input mean visible already in layer 1 (Figure~\ref{fig:cifar}d). Unlike MNIST, global standardization lowers CIFAR-10 input mean energy. Even here, no single cosine threshold separates all conditions: BP can also have high cosine on uncentered images. Table~\ref{tab:cifar_controls} supplies participation and accuracy for that reason.

\paragraph{Label-prior shift after learning.}\label{app:label_shift}
To test whether imbalance can restart collapse after learning, we return to globally standardized MNIST with a three-layer, width-300 tanh MLP and softmax cross-entropy. Training initially samples the empirical class distribution; at the shift, each example is drawn from class 0 with probability $0.7$ and from the original training distribution otherwise. The fixed probe retains the original distribution. With imbalance introduced at step 1500, the post-shift peak hidden cosine is \numShiftCos\ and balanced accuracy is \numShiftMinorityPre\ after the first shifted update (step 1500) and \numShiftMinorityPost\ at step 1750. Table~\ref{tab:E8} separates pre-shift and post-shift windows, including earlier shifts and matched BP controls. The first shifted batch has mean-error norm \numShiftEbarAtShift, falling to \numShiftEbarPost\ within 50 updates. Confident, correct predictions contribute little error regardless of class frequency. % provenance-ok: label-shift protocol

\paragraph{Calibrating an imbalanced softmax readout.}\label{app:imbcal}
Three paired runs of the same MNIST MLP use $p_0=0.7$ and standardized inputs. Setting the initial mean logits to the log sampling prior reduces the initial mean error to \numImbCalEbar\ and peak top-layer cosine from \numImbUncalCos\ to \numImbCalCos. However, balanced accuracy after 450 updates is \numImbCalBalEarly\ rather than \numImbUncalBalEarly; both approach \numImbCalBalEnd\ by update 3,000. BP also shows a modest early delay. Thus suppressing the high-cosine transient does not guarantee faster balanced-accuracy improvement under imbalance. These measurements average class accuracies; they do not separately resolve the learning trajectory of each minority class. % provenance-ok: mixture weight, horizon and checkpoint

% Generated by scripts/supplement_tables.py.
\begin{table}[t]\centering\footnotesize\setlength{\tabcolsep}{3.5pt}
\caption{CIFAR-10 controls. Conditions shared with Figure~\ref{fig:cifar} use the same runs and layers. CNN geometry uses hidden block 3, the 128-unit fully connected layer after two convolutional blocks; MLP geometry uses layer 1. Entries average per-run peak cosine over 4000 updates and minimum participation within 300 updates, with sample standard deviations. Accuracy is measured at step 3000. Softmax CNNs use globally standardized inputs. Other rows use raw pixels except the stated centering or standardization controls. Imbalanced accuracy should be read with Table~\ref{tab:review_imbalance}.}\label{tab:cifar_controls}
\begin{tabular}{lrrrr}\toprule
Condition & $n$ & Peak cosine & Min. participation & Accuracy\\\midrule
CNN: plain DFA & 3 & $1.000\pm0.000$ & $0.009\pm0.001$ & $0.276\pm0.021$\\
CNN: BP & 3 & $0.964\pm0.008$ & $0.756\pm0.026$ & $0.343\pm0.006$\\
CNN: prior bias & 3 & $1.000\pm0.000$ & $0.016\pm0.002$ & $0.349\pm0.039$\\
CNN: calibrated logits & 3 & $1.000\pm0.000$ & $0.016\pm0.010$ & $0.340\pm0.015$\\
CNN: center error & 3 & $0.876\pm0.013$ & $0.905\pm0.015$ & $0.358\pm0.031$\\
CNN: center error + inputs & 3 & $0.139\pm0.030$ & $0.850\pm0.023$ & $0.349\pm0.019$\\
CNN: balanced softmax & 3 & $0.150\pm0.025$ & $0.854\pm0.043$ & $0.431\pm0.024$\\
CNN: balanced softmax, BP & 3 & $0.054\pm0.007$ & $0.992\pm0.003$ & $0.404\pm0.011$\\
CNN: imbalanced softmax & 3 & $0.965\pm0.029$ & $0.146\pm0.090$ & $0.729\pm0.007$\\
CNN: imbalanced softmax, BP & 3 & $0.375\pm0.059$ & $0.983\pm0.004$ & $0.727\pm0.008$\\
MLP: plain DFA & 3 & $0.995\pm0.001$ & $0.111\pm0.011$ & $0.333\pm0.016$\\
MLP: BP & 3 & $0.853\pm0.009$ & $0.804\pm0.005$ & $0.335\pm0.006$\\
MLP: calibrated logits & 3 & $0.965\pm0.006$ & $0.346\pm0.063$ & $0.358\pm0.008$\\
MLP: center error & 3 & $0.805\pm0.008$ & $0.842\pm0.027$ & $0.312\pm0.021$\\
MLP: center inputs & 3 & $0.046\pm0.003$ & $0.940\pm0.017$ & $0.375\pm0.005$\\
MLP: standardized inputs & 3 & $0.159\pm0.036$ & $0.937\pm0.003$ & $0.414\pm0.006$\\
MLP: center error + inputs & 3 & $0.002\pm0.001$ & $0.933\pm0.009$ & $0.351\pm0.004$\\
\bottomrule\end{tabular}\end{table}

% Generated by scripts/review_reporting.py.
\begin{table}[t]\centering\scriptsize\setlength{\tabcolsep}{3.5pt}
\caption{Imbalanced softmax controls at their final recorded update (MNIST: 3000; CIFAR-10: 4000). The majority-class reference uses the training sampling prior; balanced accuracy averages class recall on the fixed probe. Raw accuracy alone can be close to the majority predictor.}\label{tab:review_imbalance}
\begin{tabular}{lrrrr}
\toprule
Dataset / rule & $n$ & Majority reference & Accuracy & Balanced accuracy\\\midrule
CIFAR-10, DFA & 3 & $0.73\pm0.00$ & $0.74\pm0.00$ & $0.21\pm0.03$\\
CIFAR-10, BP & 3 & $0.73\pm0.00$ & $0.73\pm0.01$ & $0.18\pm0.02$\\
CIFAR-10, centered error & 3 & $0.73\pm0.00$ & $0.71\pm0.00$ & $0.21\pm0.01$\\
MNIST, DFA & 15 & $0.73\pm0.00$ & $0.96\pm0.01$ & $0.88\pm0.01$\\
MNIST, BP & 5 & $0.73\pm0.00$ & $0.95\pm0.01$ & $0.83\pm0.01$\\
\bottomrule\end{tabular}\end{table}

% Generated by scripts/supplement_tables.py.
\begin{table}[t]\centering\footnotesize\setlength{\tabcolsep}{3.5pt}
\caption{Class-prior shifts in the MNIST softmax MLP described in Appendix~\ref{app:label_shift}. Peak cosine uses the 100 updates before and 300 updates after the shift. Error norms use the first shifted training batch and the batch 50 updates later. The fixed validation probe retains its pre-shift distribution.}\label{tab:E8}
\begin{tabular}{lrrrrrr}\toprule
Condition & $n$ & Shift & Cos. before & Cos. after & $\|\bar e\|$: first & +50\\\midrule
DFA, early shift & 3 & 300 & $0.09\pm0.07$ & $0.06\pm0.04$ & $0.473\pm0.033$ & $0.082\pm0.005$\\
BP, early shift & 3 & 300 & $0.22\pm0.01$ & $0.21\pm0.01$ & $0.414\pm0.067$ & $0.098\pm0.014$\\
DFA, late shift & 3 & 1500 & $0.03\pm0.01$ & $0.03\pm0.01$ & $0.105\pm0.019$ & $0.035\pm0.017$\\
BP, late shift & 3 & 1500 & $0.11\pm0.00$ & $0.11\pm0.00$ & $0.172\pm0.025$ & $0.069\pm0.006$\\
\bottomrule\end{tabular}\end{table}

\begin{figure*}[t]\centering\includegraphics[width=\textwidth]{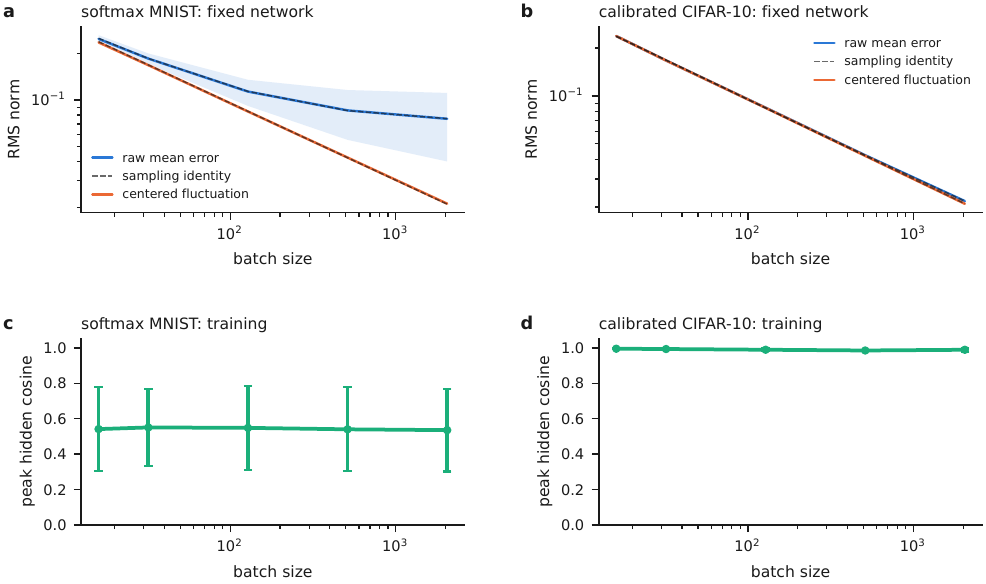}
\caption{\textbf{Sampling noise and collapse are separate measurements.} (a,b) Root-mean-square batch-mean error at frozen initialization: raw norm and fluctuation about the population mean, with the sampling identity shown dashed. (c,d) Peak top-layer cosine during 1,500 updates with the same readout choices. Panels a,c use softmax MNIST; b,d use a mean-logit-calibrated sigmoid CIFAR-10 MLP. Curves show means and bands or bars one standard deviation across three seeds. Changing batch size also changes the number of examples processed.}\label{fig:revision_batch}\end{figure*}

\subsection{Batch fluctuations and the number of classes}
\label{app:batch_test}
To separate sampling fluctuations from learning, we measure errors over the training split at fixed initialization. Let $m_e$ and $\Sigma_e$ denote their population mean and covariance, and $\bar e_n$ the error mean of a sampled batch of size $n$. The expectation below is over sampled batches at this fixed network state.

For independent sampling with replacement,
\begin{equation}
\mathbb E\|\bar e_n\|^2=\|m_e\|^2+\operatorname{tr}(\Sigma_e)/n,
\qquad \mathbb E\|\bar e_n-m_e\|^2=\operatorname{tr}(\Sigma_e)/n.
\end{equation}
We draw 512 batches at each size in $\{16,32,128,512,2048\}$, for three seeds per setting. Fitted centered-RMS exponents range from \numNoiseSlopeMin\ to \numNoiseSlopeMax, consistent with $-1/2$ (Figure~\ref{fig:revision_batch}a,b). The raw norm need not follow that law when $m_e\ne0$. % provenance-ok: batch-noise sampling protocol and theoretical exponent

We separately train all batch-size conditions for 1,500 updates (panels~c,d). MNIST has lower peak cosine, whereas calibrated CIFAR-10 stays close to one throughout the tested batch-size range. Its dependence on batch size is not monotonic. Thus the fixed-state noise identity does not establish a universal $n^{-1/2}$ collapse law or explain the CIFAR trajectory by sampling noise alone. Training also changes gates, calibration error and feedback--activity correlations; the comparison holds updates, not processed examples, fixed.

The MNIST batch sweep uses pixels divided by 255, whereas the headline softmax condition uses global standardization. Its peak cosine therefore differs from the headline value without contradicting it. Table~\ref{tab:review_heads} aligns horizons and seed pairs and includes sigmoid and softmax with the same raw-pixel preprocessing.
% Generated by scripts/review_reporting.py.
\begin{table}[t]\centering\scriptsize\setlength{\tabcolsep}{3.5pt}
\caption{MNIST readout comparisons with preprocessing stated explicitly. All rows use the same three initialization--feedback pairs, $(0,0)$, $(1,1)$ and $(2,2)$, batch size 128, and a 1,500-update window for peak cosine; participation minima use the first 300 updates. Table~\ref{tab:E1} reports larger 15-trajectory cohorts for the corresponding headline protocols; the raw-pixel softmax row is an additional preprocessing-matched comparison. Only the first two rows isolate the readout choice.}\label{tab:review_heads}
\begin{tabular}{lrrrr}
\toprule
Protocol & $n$ & Initial $\|\bar e\|$ & Peak cosine & Min. $p_3$\\\midrule
Raw pixels, sigmoid & 3 & $1.289\pm0.055$ & $0.998\pm0.000$ & $0.214\pm0.019$\\
Raw pixels, softmax & 3 & $0.078\pm0.032$ & $0.548\pm0.237$ & $0.920\pm0.045$\\
Standardized, softmax & 3 & $0.091\pm0.002$ & $0.297\pm0.024$ & $0.930\pm0.011$\\
\bottomrule\end{tabular}\end{table}

CIFAR-100 provides a larger-class-count check with the same three-layer tanh MLP. Softmax starts with much smaller mean error than sigmoid outputs, yet both readouts produce high-cosine transients; gate concentration is substantially milder with softmax (Table~\ref{tab:cifar100}). Matched BP also has high cosine on these uncentered images, but DFA concentrates gate energy far more strongly under both readouts and learns more slowly; after 3,000 updates both rules are still early in training. The ideal sigmoid formula is $\sqrt C(1/2-1/C)$. Uniform softmax predictions give zero mean error for a balanced class prior; more generally, mean error vanishes whenever the mean prediction matches the target prior. These runs change dataset as well as class count, so they are generality checks rather than an isolated class-count test or competitive benchmark.
% Generated by scripts/supplement_tables.py.
\begin{table}[t]\centering\footnotesize\setlength{\tabcolsep}{3.5pt}
\caption{CIFAR-100 readouts in the three-layer tanh MLP under DFA and matched BP (same initializations and minibatches). High peak cosine also occurs under BP on these uncentered images; participation separates the rules, and sigmoid outputs concentrate gate energy most. Participation minima use the first 300 updates; peak cosine and final accuracy use the 3000-update run. Accuracy remains low for both rules after 3000 updates at this learning rate, so these runs test the early transient rather than final performance.}\label{tab:cifar100}
\begin{tabular}{lrrrrr}\toprule
Readout, rule & $n$ & Initial $\|\bar e\|$ & Peak cosine & Min. $p_3$ & Accuracy\\\midrule
Sigmoid, DFA & 3 & $4.976\pm0.015$ & $1.000\pm0.000$ & $0.036\pm0.007$ & $0.011\pm0.004$\\
Sigmoid, BP & 3 & $4.976\pm0.015$ & $0.999\pm0.000$ & $0.282\pm0.006$ & $0.018\pm0.005$\\
Softmax, DFA & 3 & $0.046\pm0.004$ & $0.986\pm0.011$ & $0.632\pm0.100$ & $0.028\pm0.004$\\
Softmax, BP & 3 & $0.046\pm0.004$ & $0.777\pm0.020$ & $0.954\pm0.003$ & $0.069\pm0.007$\\
\bottomrule\end{tabular}\end{table}

\clearpage
\begin{figure*}[t]\centering\includegraphics[width=\textwidth]{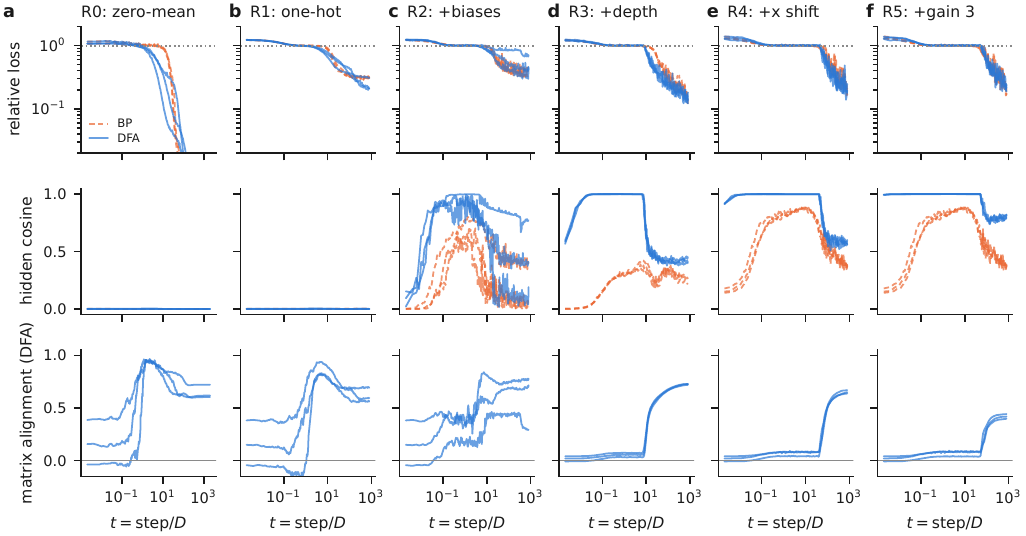}
\caption{\textbf{The teacher--student ladder.} Test loss relative to the constant predictor (top), top-layer hidden cosine (middle) and DFA weight alignment (bottom) against $t=\mathrm{step}/D$, where $D$ is input dimension. DFA is solid and BP dashed; thin lines show the three individual seeds. Column a adapts \citet{refinetti2021} to four outputs; b--f change the objective and targets, hidden biases, depth and activation, input statistics, then feedback gain. Weight alignment is the cosine between vectorized feedback and downstream weight matrices (Appendix~\ref{app:diagnostics}).}\label{fig:ladder}\end{figure*}

\subsection{Comparison with align-then-memorise dynamics}
\label{app:demarcation}
The finite-width saturation transient complements analyses of lazy and feature-learning regimes \citep{bordelon2023}. It also differs from class-structured neural collapse \citep{papyan2020}, self-supervised dimensional collapse \citep{jing2022}, and previously studied BP plateaus \citep{fukumizu2000,amari2018,saxe2014}. Saturation need not always hinder learning: \citet{frenkel2021} exploit fixed random target projections interacting with saturating units.

A plateau in loss can have different mechanisms across learning problems. Figure~\ref{fig:ladder} compares our stall with an adaptation of the scalar-regression setting of \citet{refinetti2021}: Gaussian inputs, error-function (erf) hidden units, zero-mean vector regression targets and online SGD, with a teacher of $K=4$ hidden units, a student of $M=4$, four outputs and input dimension $D=500$. Column b switches to one-hot classification targets and sigmoid cross-entropy; c adds trainable hidden biases; d uses three tanh layers of width 300; e shifts and rescales inputs; and f raises feedback gain to $3$. These are protocol comparisons, not single-factor interventions in every column. % provenance-ok: ladder architecture and panel settings

The experiment uses three seeds per condition, fresh Gaussian samples at every step, hidden learning rate $0.5$ and output rate one tenth as large; horizontal axes show updates divided by $D$. Plateau windows for the synthetic runs are detected with a $\pm5\%$ band around the constant-predictor loss and no accuracy criterion, unlike the image-classification detector. Both window summaries are retained in the analysis outputs (Appendix~\ref{app:code}). % provenance-ok: ladder protocol

\numLadderVerdict. In rungs R3--R5, BP also stays near the constant predictor for a similar time without full collapse, so in this synthetic task collapse does not set the plateau length. In these controls, nonzero-mean targets supply a mean error and trainable biases create an additional mean-drive channel. A flat loss near the constant predictor need not involve collapse: the cosine and alignment rows distinguish a pre-alignment plateau with hidden cosine near one from the post-alignment plateau in the regression control, which is reached at high weight alignment with the cosine low; the two can also coexist within one trajectory. This comparison is empirical, not an if-and-only-if consequence of Proposition~\ref{prop:rankone}.

\end{document}